\pdfoutput=1
\documentclass{IEEEtran}
\usepackage{cite}
\usepackage{amsmath, amssymb, amsfonts}
\usepackage{algorithmic}
\usepackage{graphicx}
\usepackage{textcomp}
\usepackage{color}
\usepackage{graphicx}
\usepackage{mathrsfs}
\usepackage{mdwtab}
\usepackage{eqparbox}
\usepackage[tight, footnotesize]{subfigure}
\usepackage{stfloats}

\usepackage{booktabs}
\usepackage[ruled, vlined]{algorithm2e}
\usepackage{enumerate}
\usepackage{amsthm}
\usepackage{amsmath}
\usepackage{tikz}
\usepackage{adjustbox}
\usepackage{pgfplots}
\usepackage{longtable}
\usetikzlibrary{shapes}
\usepackage{multicol}

\newsavebox{\mysavebox}

\def\be{\boldsymbol e}

\def\BibTeX{{\rm B\kern-. 05em{\sc i\kern-. 025em b}\kern-. 08em
    T\kern-. 1667em\lower. 7ex\hbox{E}\kern-. 125emX}}
\begin{document}
\title{Robust Key-Frame Stereo Visual SLAM with low-threshold Point and Line Features}
\author{Meiyu Zhi

\thanks{This work was supported in part by the National Natural Science Foundation of  China under Grants  (62121004,  61876041),   the Local Innovative and Research Teams Project of Guangdong Special Support  Program (2019BT02X353),  Key Area Research and Development Program of Guangdong Province (2021B0101410005),  and Guangdong Basic and Applied Basic Research Foundation (2021B1515420008). }% <-this % stops a space
\thanks{H. Rao and Y. Xu are with the Provincial Key Laboratory of Intelligent Decision and Cooperative Control,  School of Automation,  Guangdong University of Technology,  Guangzhou 510006,  China. (e-mail: raohxia@163.com, xuyong809@163. com)}% <-this % stops a space

}

% \markboth{IEEE Transactions on Automatic Control}%
% {Shell \MakeLowercase{\textit{et al. }}: Bare Demo of IEEEtran. cls for IEEE Journals}
\maketitle

\begin{abstract}
In this paper, we develop a robust, efficient visual SLAM system that utilizes spatial inhibition of low threshold, baseline lines, and closed-loop keyframe features. Using ORB-SLAM2, our methods include stereo matching, frame tracking, local bundle adjustment, and line and point global bundle adjustment. In particular, we contribute re-projection in line with the baseline. Fusing lines in the system consume colossal time, and we reduce the time from distributing points to utilizing spatial suppression of feature points. In addition, low threshold key points can be more effective in dealing with low textures. In order to overcome Tracking keyframe redundant problems, an efficient and robust closed-loop tracking key frame is proposed. The proposed SLAM has been extensively tested in KITTI and EuRoC datasets, demonstrating that the proposed system is superior to state-of-the-art methods in various scenarios.
\end{abstract}

\begin{figure}[htbp]
\centering
    \subfigure[Point and line features are witnessed in one image.
.]{
    \begin{minipage}[t]{\linewidth}
    \centering
    \includegraphics[width=8 cm]{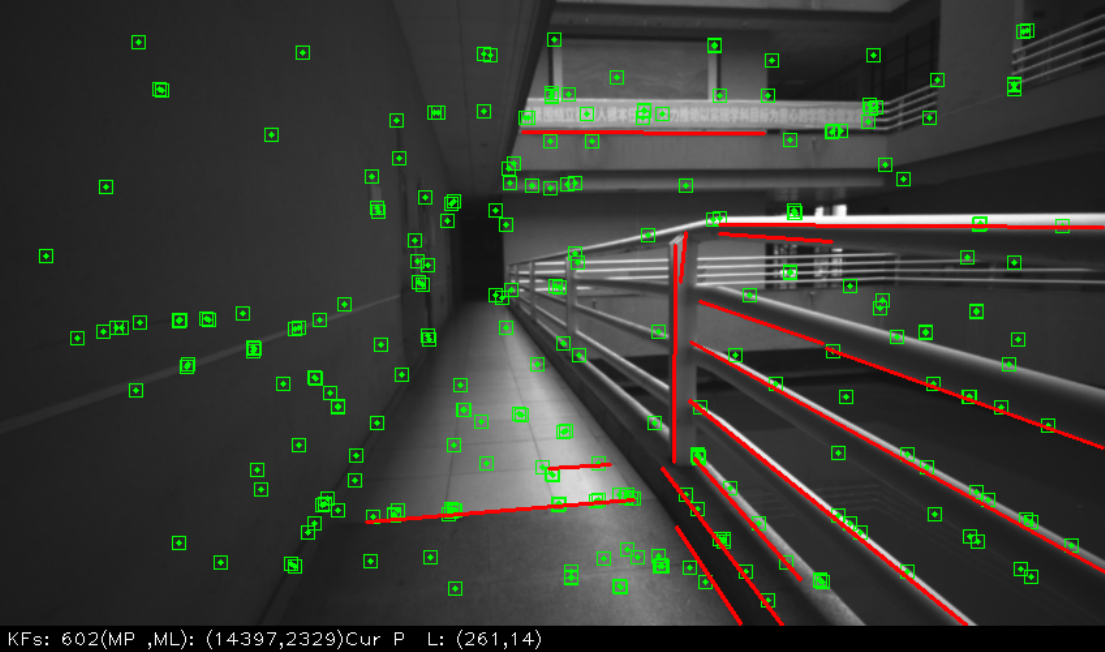}
    \end{minipage}%
    }%
    \\
    \subfigure[The point and line map]{
    \begin{minipage}[t]{\linewidth}
    \centering
    \includegraphics[width=8 cm]{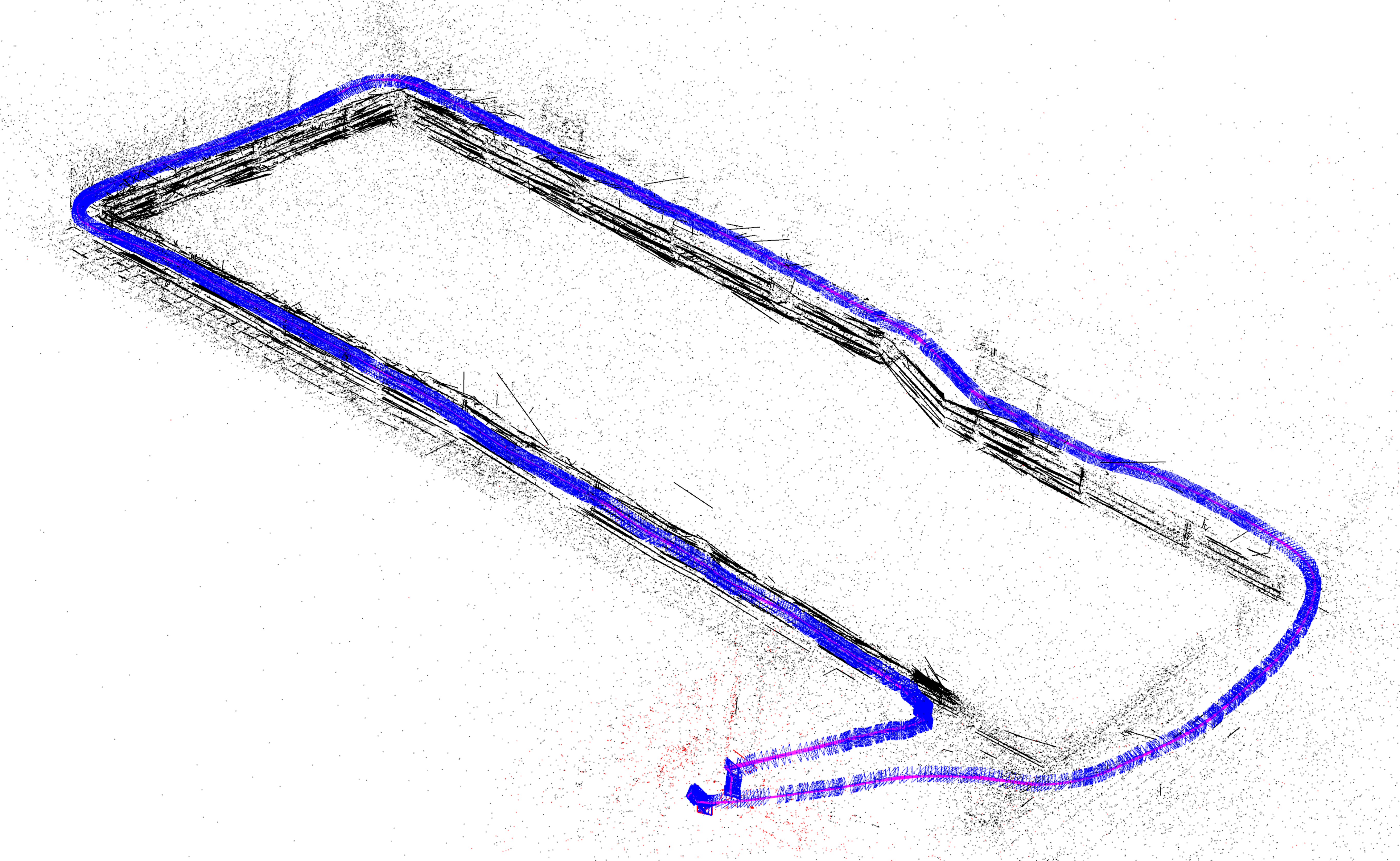}
    \end{minipage}%
    }%

\centering
\caption{The proposed visual SLAM with point and line features on our dataset. Note that in (b),   the purple lines indicate the trajectory of camera motion. The blue frames represent keyframes,   the current frame in red,   and the local map for the Tracking at that moment in red.}
\label{Introduction}
\end{figure}
\section{Introduction}
\IEEEPARstart{V}{isual} Localization and Mapping algorithms are used to estimate the 6D camera pose while recreating unknown environments, which are critical for autonomous robots and augmented reality because camera attitude estimation enables cars and Unmanned aerial vehicles to position themselves. 
\par VSLAM is divided into two direct methods: DSO\cite{wang2017stereo} and SVO\cite{forster2014svo}. and feature point methods such as ORB-SLAM2\cite{mur2017orb}. The feature point method uses continuous frame tracking to identify key points and then recovers camera motion trajectory and 3-D points. ORB-SLAM2 is considered to be the current state-of-the-art SLAM system. It is developed based on many excellent works. e. g. , PL-SLAM\cite{pumarola2017pl, gomez2019pl}, an efficient graph-based bundle adjustment (BA) algorithm. The point-based approach is robust because the key points are invariant relative to views and illumination. Therefore, The VSLAM system screening key points to maintain robustness. The quad-tree\cite{mur2017orb} and kd-tree\cite{buoncompagni2015saliency} are important data structures for screening key points to solve key points redundancy. In addition, low threshold keypoints can be more effective in dealing with low textures.   \par

Recently, combined line features and point features have become favorable for the VSLAM due to low texture and adaptation to a more realistic environment such as indoor and outdoor scenes. Dealing with low texture-specific SLAM methods based on lines and points, such as S-SLAM\cite{li2020structure}, Stereo-PL-SLAM \cite{gomez2019pl,  zuo2017robust} For reconstruction, point and line SLAM is sparse mapping. Compared to point SLAM, it has only an incomplete map. Having lines in maps improves the approximation accuracy and has 3D reconstruction on line\cite{li2021rgb}. However, small errors accumulate over time. To overcome these errors, strategies are loop-closure detection\cite{strasdat2010scale} and reconstruct a high accuracy map\cite{zuo2017robust}. Loop closure detection, combined with pose graph optimization, detects previously seen landmarks and optimizes the pose graph based on new constraints. However, Loop closure adds an extra computational burden and removes the drift only when revisiting the same place. Another strategy is to build a global landmark in the world frame. Extracting lines by Line Segment Detector(LSD)\cite{von2012lsd}contain broke lines. Therefore, the connection disconnection can effectively improve localization precision\cite{huang2020lidar}.  \par

The keyframe strategy has high accuracy for localization\cite{mur2017orb}. Keyframe extraction reduces the number of images to be processed and reduces redundant information\cite{jiang2016human}. An appropriate number of keyframes is critical for stability. \par

In this paper, We build on our Stereo-SLAM and propose a robust SLAM designed to deal with and adapt to various environments, including low texture, reducing time, and improving tracking and mapping simultaneously. Fig .\ref{Introduction} presents a tracking and mapping process containing points and line segments. Different to \cite{mur2017orb, gomez2019pl, wang2017stereo}, the baseline is merged into error line functions and line Jacobian, which improves the estimation accuracy. In summary, our contributions are:
% 无序列表环境:
\begin{itemize}
    \item An improved extraction method for low-threshold points is introduced to reduce the time of the program.
    \item An improved extraction method for connecting broke lines is introduced to robustify data association. 
    \item In the back-end of the proposed visual SLAM, we employ the baseline to calculate line functions and line Jacobians, which apply to frame tracking, local mapping, and global bundle adjustment. 
    \item We design a robust keyframe strategy in which we could reasonably add keyframes to keep the system strong. 
\end{itemize}

\section{RELATED WORK}\label{sec:pro}
% \subsection{Point Visual SLAM}
PTAM\cite{klein2009Parallel} is a monocular, keyframe-based SLAM system that was the first multithreaded system to include tracking and mapping and has been successful in real-time. As a state-of-the-art SLAM system, ORB-SLAM2 combines feature-based Tracking, sparse point Mapping, and loop closure with the Bag-of-words model. It needs to add keyframes by observation inliers between the reference frame and current frame, decreasing the quality of tracking ORB-SLAM2 insert keyframes. This method FastORBSLAM\cite{fu2020fastorb} replaces ORB-SLAM2 descriptors Kanade-Lucas-Tomasi Tracking Method(KLT), which reduces match points-to-points time. Oleksandr Bailos'\cite{bailo2018efficient} proposes an efficient adaptive non-maximal suppression algorithm for keypoints distribution to minimize the time to finish the program. This algorithm finds a global response value more significant than the set threshold. LSD\cite{von2012lsd} is a short line segment detection to adopt a pyramid of images to determine the picture's location. Compared toTHRESH-OTSU to extract the line, LSD extracted lines more accurately. LBD(Line Band Descriptor)\cite{zhang2013efficient} is a method describing a line of feature using a binary string which is an efficient method to match a line. XOR operations applying to Binary string improved match line-to-line.  \par
 Inspired by the LSD and LBD, PL-SLAM\cite{pumarola2017pl} incorporates lines which include frame tracking, local mapping, bag-of-words about line, and global BA. In addition, PL-SLAM in texture with fewer points can run steady. RPL-SLAM\cite{zuo2017robust} employs the orthonormal representation to parameterize lines and analytically compute the Jacobians, which is the first system to operate the orthonormal re-presentation as the minimal parameterize to model lines. BowPL\cite{qian2019visual} merge ORB-SLAM2 with PL-SLAM with the Bag-of-words model. To connect broken lines, RPL-SLAM connects lines by points-to-lines distance to improve estimate accuracy. Recent representative works include semi-direct methods about lines and sparse direct methods about the line. Gomez-Ojeda. Propose semi-direct visual odometry(PL-SVO)\cite{gomez2016pl}, a two-thread framework that consists of Tracking and Local Mapping with line. It tracks sparse pixels 
at the FAST corners and lines at LSD to recover motion in Tracking and reﬁnes the pose in Local Mapping. SVO uses a depth ﬁlter model to estimate pixel depth values and ﬁlters outliers\par
 DSO\cite{wang2017stereo} is the direct method, keyframe SLAM system by the pixel's gray value. Regardless of changes in optical flow and exposure parameters under rotation,  a new keyframe is created when the weighted sum of optic flow and exposure parameters is greater than 1. The new Keyframe will be used for subsequent sliding window optimization. LSD-SLAM\cite{engel2014lsd} is a large-scale direct monocular SLAM based on Keyframe by change of position. 

\section{SYSTEM OVERVIEW}\label{sec:main}
Our approach visual SLAM system is based on ORB-SLAM2 and has two different parallel threads see(Fig. 2): Tracking, Local Mapping. Lines are not used to detecting loop place but for global BA. In the following,  we briefly describe each component while focusing on the difference between ORB-SLAM2. 
\begin{figure}\textbf{}
  \centering
  \includegraphics[width=8cm]{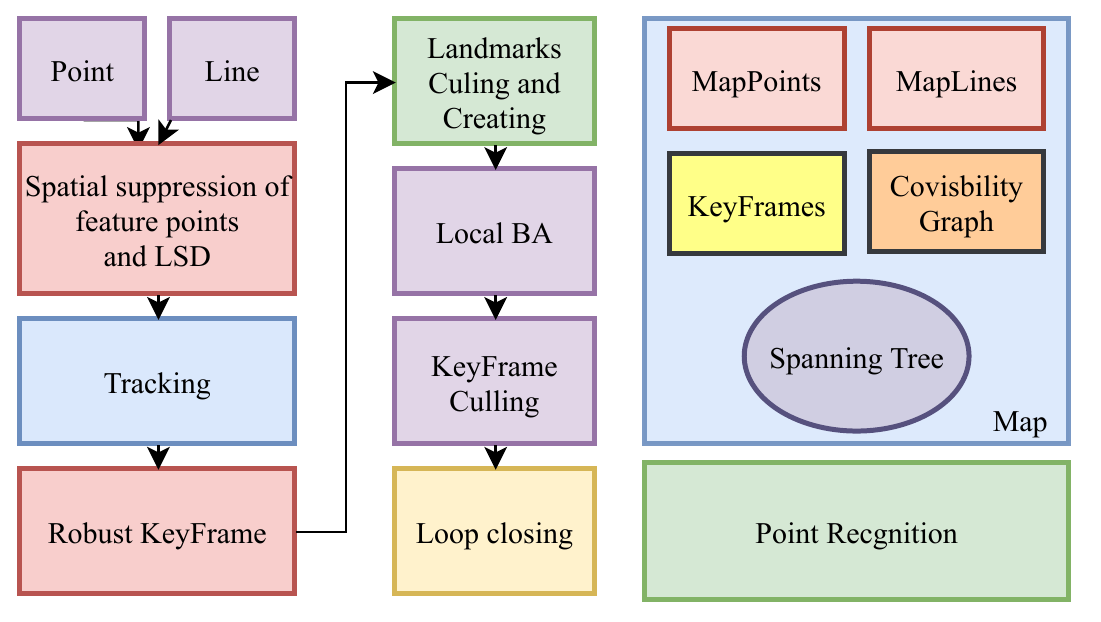}\\
  \caption{The architecture of the proposed graph-based visual SLAM system uses fast point, line, and Robust KeyFrame.}
\end{figure}

\subsection{Spatial suppression of feature points}
\begin{figure}\textbf{}
  \centering
  \includegraphics[width=8cm]{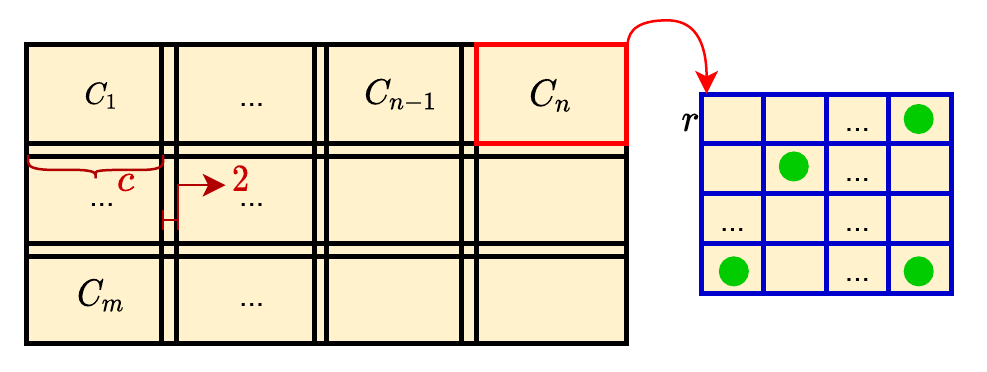}\\
  \caption{Feature points were extracted via Spatial suppression with 64 trees. }
\end{figure}
As shown in Fig.  3,  The image consists of multiple cells. The length of the side is set $c$ in each cell. Each cell keeps $2$ pixels and has 64 nodes. The Cols of the image are $W$, and The rows of the image are $H$. 
Assuming the number of reasonable feature points in this image pyramid is $N$. $n$ is the number of cols of points,  and $m$ is the number of rows of points. Then, considering image resolution, demonstrate the spatial relationship between cols points, rows points, and cells of image computed as follows: 
\begin{equation}\label{eq:1-1}
\left\{\begin{aligned}
  W&=c+(n-1)(\frac{c}{2}+1)\\
  H&=c+(m-1)(\frac{c}{2}+1), 
\end{aligned}
\right. 
\end{equation}
where implies that $N=m \times n$, which the length of side $c$ in each cell is equal to
\begin{equation}\label{1-2}
\left\{\begin{aligned}
    \delta &=16(N-1)(W+H+HW-N+1)\\
    \Delta &=\sqrt{(W+2H+2N+2)^2+\delta}\\
    c &=\frac{-2(W+2H+2N+2) \pm ± \Delta}{2(N-1)} \quad (c \geq 64). 
\end{aligned}
\right. 
\end{equation}
Restraining the width of the $r$ is used to reduce the number of redundant feature points and is expressed as(\ref{1-3}). Only one of most responses for points is an inlier with this width,  which we choose to use. Now we have a 64 tree about the length of side $c$. 
\begin{equation}\label{1-3}
\begin{split}
    r=\frac{c}{8}
\end{split}
\end{equation}
\subsection{ Extraction of Line Features}
Line Segment detector (LSD) is a commonly used feature to describe line segment texture and is a fast detection method. However, the LSD suffers from the problem of dividing a line into multiple segments\cite{zuo2017robust}. Therefore, this paper attempts to improve the LSD algorithm by mapping image points to Hough space. The center of the image resolution is an origin divided into four quadrants. Therefore, a line $l$ with startpoint$l(x_1, y_1)$ and endpoint$l(x_2, y_2)$ . The line of the slope is $k$, and the line of translation is $b$ as follows: 
\begin{equation}\label{1-4}
\left\{\begin{aligned}
    k &=\frac{y_1-y_2}{x_1-x_2}\\
    b &=y_1-kx_1. 
\end{aligned}
\right. 
\end{equation}
From the line in the Hough space, the line is mapped as $l(\rho, \theta)$. The procedure can be expressed as
\begin{equation}\label{1-5}
\left\{\begin{aligned}
    \theta &=-\arctan \frac{1}{k},  \quad (k<0, b>0)\\
    \theta &=\pi -\arctan \frac{1}{k}, \quad (k>0, b>0),\\
    \theta &=\pi -\arctan \frac{1}{k}, \quad (k<0, b<0),\\
    \theta &=2\pi -\arctan \frac{1}{k}, \quad (k>0, b<0),\\
    \rho &=x_1\cos{\theta}+y_1\sin{\theta}. 
\end{aligned}
\right. 
\end{equation}
In our approach, we merge the segments according to the same point in Hough space. If $\rho$ smaller than similar line $1\%$ and $ \theta < \frac{\pi}{180} $, we fit the straight line through least-squares method with four points. The procedure of fusion is shown in Fig. 4. As our experiments demonstrate, this improved line detector has the advantage of making data associations more robust and accurate. Note that the merged line segments found by our improved detector is represented by an the LBD line descriptor, which is a 256-bit vector, the same as the ORB point descriptor. 
\begin{figure}[htbp]
\centering
    \subfigure[Lines are extracted from LSD.]{
    \begin{minipage}[t]{0.48\linewidth}
    \centering
    \includegraphics[width=4 cm]{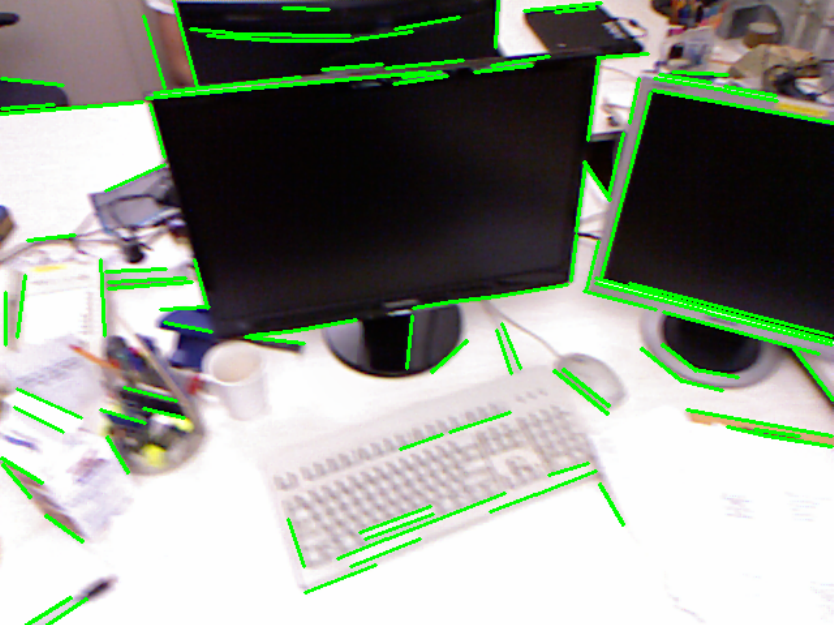}
    %\caption{fig1}
    \end{minipage}%
    }%
    \subfigure[Suitable for merge Lines are merged]{
    \begin{minipage}[t]{0.48\linewidth}
    \centering
    \includegraphics[width=4 cm]{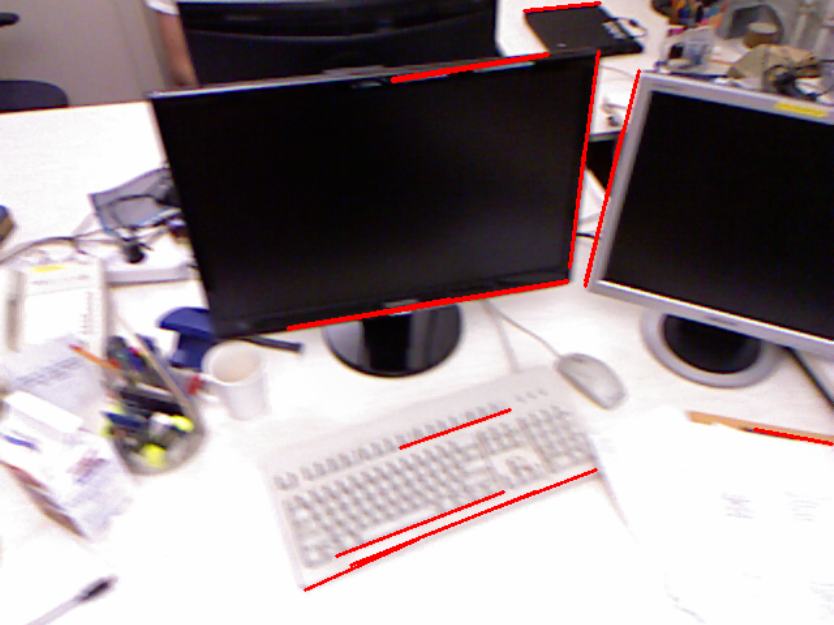}
    %\caption{fig2}
    \end{minipage}%
    }%
    \hspace{0.1 pt}
    \subfigure[Match for lines after merge in our approach.]{
    \begin{minipage}[t]{\linewidth}
    \centering
    \includegraphics[width=8 cm]{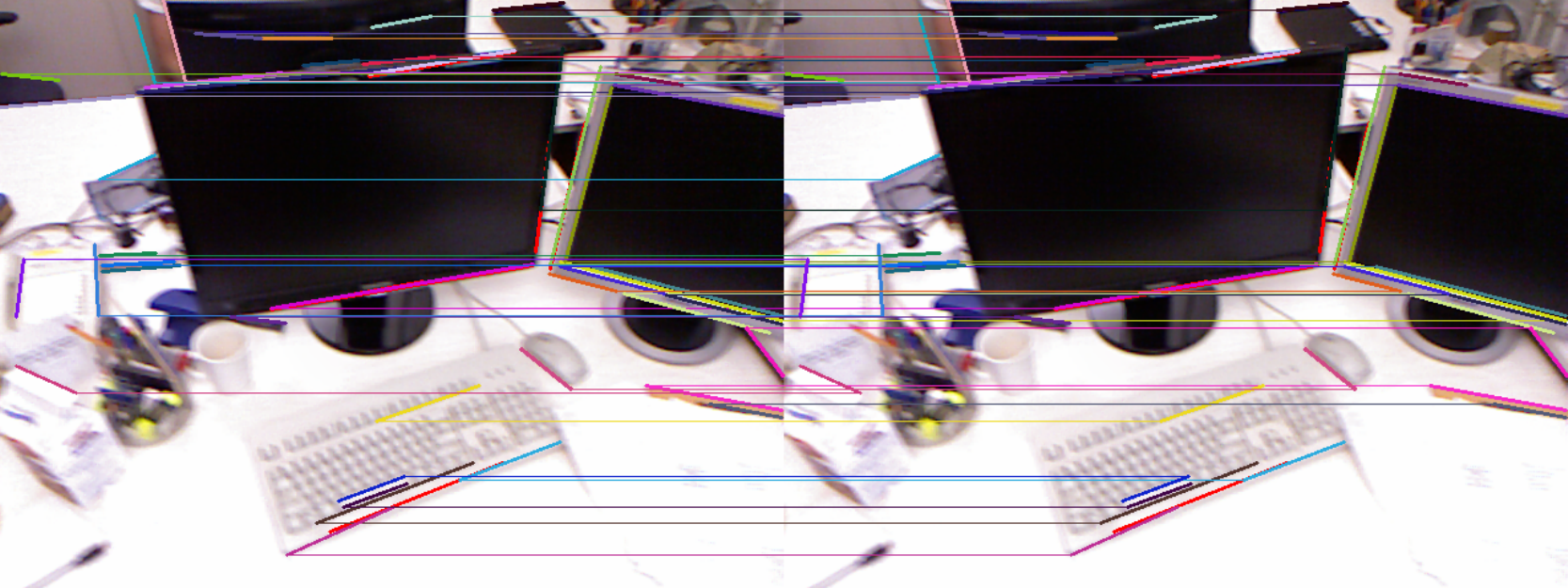}
    %\caption{fig2}
    \end{minipage}
    }%
\centering
\caption{Results of merge detectors and match.}
\end{figure}

\subsection{Jacobian of Line Re-projection Error for stereo cameras}
Given the extraction of line features, we obtain lines after fitting. we obtain the 2D points $p_s$ and $p_e$ mapping the normalized coordinate system to calculate line function $l$ as follow: 
\begin{equation}\label{1-6}
\begin{aligned}
    l &=\frac{p_s \times p_e}{||p_s||||p_e||}. 
\end{aligned}
\end{equation}
\begin{figure}\textbf{}
  \centering
  \includegraphics[width=8cm]{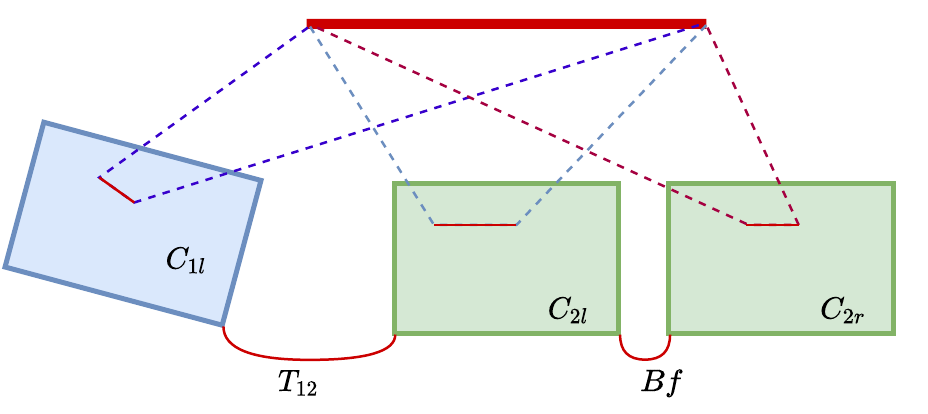}\\
  \caption{Line Re-projection Error for stereo cameras to use baseline.}
  \label{line reprojection}
\end{figure}
Then, we formulate the error function based on the point-to-line distance between $l$ and $P_l$  from the matched line. Fig .\ref{line reprojection} shows the line in the world was observed in three views. The current frame or current keyframe $C_{2l}$ and lats frame or last keyframe$C_{1l}$ present transformation $T_{12}$. And $C_{2l}$ in left camera for $C_{2r}$ present translation is baseline $Bf$. Therefore, for each 3D line, the error function can be noted as
\begin{equation}\label{1-7}
\begin{aligned}
  \be_{k, P}^l &=l \Pi(R_{k, j}P_l+t_{k, j}). 
  \end{aligned}
\end{equation}
In order to obtain the minimum optimization parameters, the re-projection error of the 3-D line is expressed as the distance between four homogeneous endpoints.
re-projection in left image are $P_{sl}(x_{sl}, y_{sl}, z_{sl})$ and $P_{el}(x_{el}, y_{el}, z_{el})$. right re-projection to left image are 
$p_{sr}(U_{sr}$, $\frac{y_{sl}f_y+c_y}{z_{sl}}, 1)$ and $p_{er}(U_{er}$, $\frac{y_{el}f_y+c_y}{z_{el}}, 1)$ of the matched line segment to the back-projected on image plane as shown in (\ref{1-8}). 
\begin{equation}\label{1-8}
\begin{aligned}
  \be=\begin{bmatrix}
   \frac{x_{sl}f_xl_x/z_{sl}+y_{sl}f_yl_y/z_{sl}+l_z}{\sqrt{l_x^2+l_y^2}}
   + \frac{x_{el}f_xl_x/z_{el}+y_{el}f_yl_y/z_{el}+l_z}{\sqrt{l_x^2+l_y^2}}\\
  \frac{(U_{sr}+bf/z_{sr})l_x+y_{sr}f_yl_y/z_{sr}+l_z}{\sqrt{l_x^2+l_y^2}}
    \\ + \frac{(U_{er}+bf/z_{er})l_x+y_{er}f_yl_y/z_{er}+l_z}{\sqrt{l_x^2+l_y^2}}
  \end{bmatrix}
  \end{aligned}
\end{equation}
Re-project the 3-D point into the current frame, and define the error function based on the re-projection, as shown in the following formula: (\ref{1-9}). 
\begin{equation}\label{1-9}
\begin{aligned}
  \be_{k, j}^p &=p_k - \Pi(R_{k, j}P_j+t_{k, j})
  \end{aligned}
\end{equation}
For our method, observations must follow a Gaussian distribution and be independent. The final non-linear least-squares cost $T^*$ can be written as in
\begin{equation}\label{1-10}
\begin{aligned}
  T^* = & argmin \sum_{j}^M \rho_p(e{_{k, j}^p}^T \Sigma_{pk, j}e_{k, j}^p)\\
  &+\rho_l(e{_{k, P_l}^l}^T \Sigma_{pk, P_l}e_{k, j}^l), 
  \end{aligned}
\end{equation}
where $\Sigma_{pk, j}$, $\Sigma_{pk, P}$ are the inverse covariance matrices of points, lines, and $\rho_p$, $\rho_l$, are robust Huber cost functions, respectively. 

Here, a solution is determined using the Levenberg Marquardt algorithm. It is known that the Jacobian is important when using an iterative approach to solve the graph optimization problem. 

In view of (\ref{1-8}), we have a error function about the small pose changes $\frac{\partial e}{\partial \delta \zeta}$ as follow: 
\begin{equation}\label{1-11}
\begin{aligned}
    \frac{\partial e}{\partial \delta \zeta}=\frac{\partial e}{\partial l_c}\frac{\partial lc}{\partial \delta \zeta}.
  \end{aligned}
\end{equation}
The re-projection error about the lines of cameras pinhole model $\frac{\partial e}{\partial l_c}$. 
\begin{equation}\label{1-12}
  \frac{\partial e}{\partial l_c} =\begin{bmatrix}
  \frac{f_xlx}{z\sqrt{l_x^2+l_y^2}} & \frac{f_yly}{z\sqrt{l_x^2+l_y^2}} &-\frac{xf_yly+yf_yl_y}{z^2\sqrt{l_x^2+l_y^2}}\\
  0 & \frac{f_yl_y}{z\sqrt{l_x^2+l_y^2}} & -\frac{bfl_x+yf_yl_y}{z^2\sqrt{l_x^2+l_y^2}}\\
  0 & 0 & 0
  \end{bmatrix}. 
\end{equation}
Not only do we optimize the poses of the system, but we also optimize landmarks. Therefore, the re-projection error about the lines of landmarks $\frac{\partial e}{\partial P_l}$ can be expressed as (\ref{1-13}), where have a rotation matrix $\textbf{R} \in SO(3)$. 
\begin{equation}\label{1-13}
\begin{aligned}
    \frac{\partial e}{\partial P_l}=\frac{\partial e}{\partial l_c} \textbf{R}
  \end{aligned}
\end{equation}
Finally, $\frac{\partial lc}{\partial \delta \zeta}$ is a jacobian about the small pose changes. The Jacobian of the re-projection error for the small pose changes and landmarks with line parameters can be written as (\ref{1-14}) and (\ref{1-15}), respectively.
\begin{equation}\label{1-14}
  j_{\zeta}
%   \frac{\partial e}{\partial \delta \zeta} 
  =
  \frac{\partial e}{\partial l_c}\frac{\partial lc}{\partial \delta \zeta}
\end{equation}
\begin{equation}\label{1-15}
  j_{P}=
%   \frac{\partial e}{\partial l_c}\textbf{R}
  = 
\begin{bmatrix}
  \frac{f_xlx}{z\sqrt{l_x^2+l_y^2}} & \frac{f_yly}{z\sqrt{l_x^2+l_y^2}} &-\frac{xf_yly+yf_yl_y}{z^2\sqrt{l_x^2+l_y^2}}\\
  0 & \frac{f_yl_y}{z\sqrt{l_x^2+l_y^2}} & -\frac{bfl_x+yf_yl_y}{z^2\sqrt{l_x^2+l_y^2}}\\
  0 & 0 & 0
  \end{bmatrix} \textbf{R}
\end{equation}

\subsection{Robust Keyframe}
\begin{figure}[htbp]
\centering
    \includegraphics[width=9 cm]{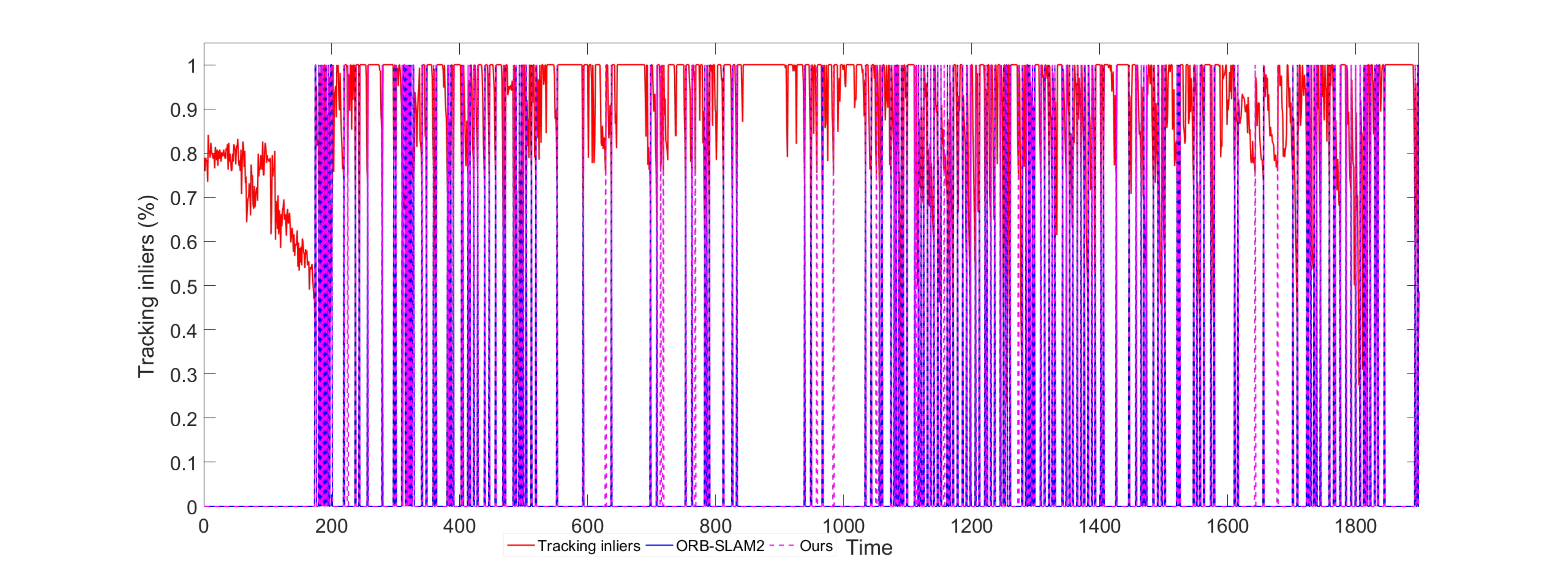}
\centering
\caption{Comparative results of different methods for selecting keyframes and selecting keyframes in EuRoC V103.}
\label{kfimage}
\end{figure}

In this way, Robust Keyframe can reduce redundant information. Keyframes below the calculated tracking values between curframe and Keyframe by a certain ratio can be expressed as (\ref{1-16}) added. However, the Keyframe of ORB-SLAM2 is poor robustness. we solve the problems to add a offset $\Delta_{PID}$ by using $PID$ as Fig .\ref{kfimage}.

\begin{equation}\label{1-16}
\begin{aligned}
  KF_{cur}<(KF_{Ref}+\Delta_{PID})*Ratio
  \end{aligned}. 
\end{equation}

If the point is very far away, the estimation of the position can lead to inaccurate results. The positions between the reference of Keyframe and current Keyframe are used to estimate velocity by way of Kalman filtering.
% see section \ref{sec:APPENDIX}. 

\section{EXPERIMENTAL RESULTS }\label{sec:simu}
\noindent
\textbf{System Implementation }
The Proposed visual SLAM system is based on ORB-SLAM2. Therefore, we describe the main implementation details of my method. 
\begin{enumerate}[1)]
\item  \emph{Spatial suppression of feature points: }
Our system uses a stereo image sequence as input. We use the cv::FAST function for every input frame to extract original features. To eliminate low texture, the threshold was used $7$ in function. We set the tolerance value of feature points $N$ to be $0. 001$. We will return all features to the used estimate state for the setting features more significant than the original features. 
\item \emph{Line Features: }
Four threads are launched to extract point features and line features. Line features are detected by LSD and described by the LBD descriptor. The LSD lines are extracted in zero octaves. Then two threads are launched for stereo matching about lines and points. The endpoints of lines to function are calculated by (\ref{1-6}). Then depth values are calculated. Therefore, Matrix information of lines is $0. 5$ in the Levenberg Marquardt algorithm.  
\item \emph{Robust Keyframe: }
Our system needs to insert Keyframe by $PID$. Error in $PID$ is calculated between the inliers of the current frame and inliers of the reference frame. $Ratio$ in our method is set to $0. 75$. 
\end{enumerate}
All experiments were carried out with an AMD Ryzen R7-5800H CPU(with @3. 2GHz), NVIDIA GeForce RTX3060 GPU and ubuntu 20. 04. We run each sequence five times and show median results for the accuracy of the estimated trajectory. We evaluate our proposed SLAM system on public datasets and compare its performances with other state-of-the-art methods. The evaluation metrics used in the experiments are the absolute trajectory error ($ATE$) and the relative pose error ($RPE$), which measure the absolute and relative pose differences between the estimated and the ground truth motion. The evaluation metrics
also include mean tracking time and feature points distributing time. \par
\noindent
\textbf{Evaluation and datasets }
In order to evaluate our method, we compare it against several stereo SLAM frameworks, as ORB-SLAM2 are state-of-the-art methods. We have also tried to compare our method against Gomez-Ojedas' PL-SLAM\cite{gomez2019pl}, but unfortunately, their approach can not run on our device. To compare our process, we compare our method against another version of Qian's PL-SLAM. This version is open source and similar to Rubens PL-SLAM. What is more, it is superior to Rubens PL-SLAM. But some datasets also can not run. \par
In the following, we present the results of Our approach compared to Qians' PL-SLAM\cite{qian2019visual} and ORB-SLAM2 in public datasets such as KITTI\cite{Geiger2012CVPR} and EuRoC\cite{Burri25012016}. Therefore, in this section, PL-SLAM indicates Qians' PL-SLAM. 
\subsection{KeyPoints allocation for time}
We evaluate the performance of times in each image pyramid. After the program end, we calculate the time average of all the allocation in each image pyramid. For a fair comparison, in our approach, we set the tolerance value to $0. 001$, and cv::Fast function for threshold both set $7$. Considering the keypoints (ORB) are extracted in an image pyramid with $L=8$ with a scale ratio of $1. 2$, our approach is implemented using the same pyramid. 
\par
\begin{figure}[htbp]
\centering
    \subfigure[The original FAST keypoints. ]{
    \begin{minipage}[t]{\linewidth}
    \centering
        \includegraphics[width=8 cm]{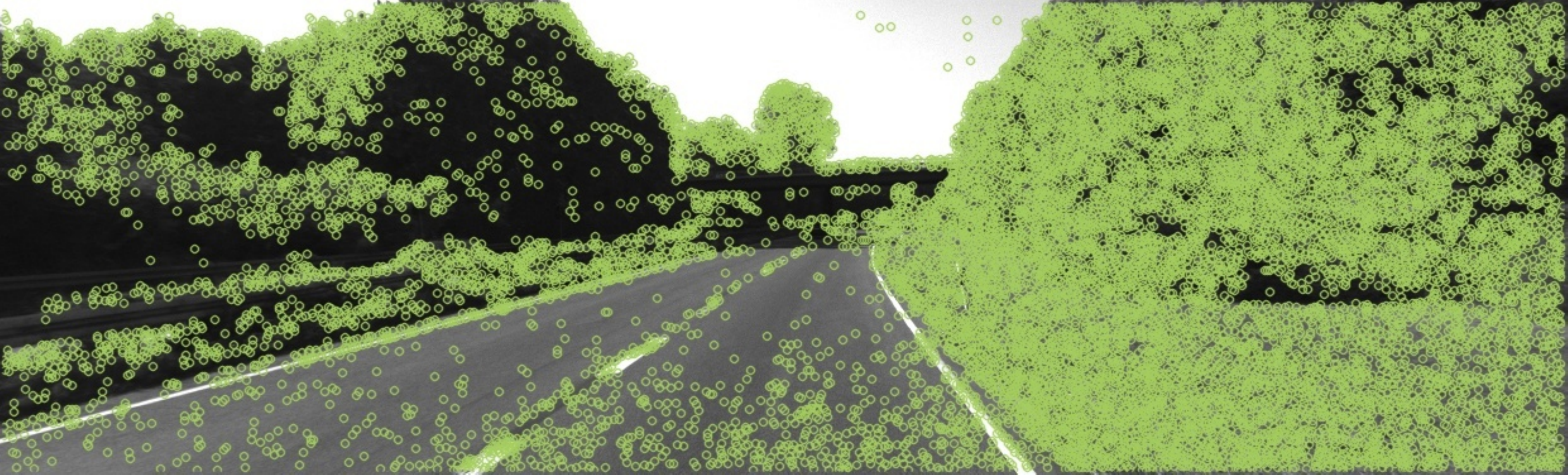}
    \end{minipage}%
    }%
    \\
    \subfigure[ORB-SLAM2 FAST keypoints of distribution. ]{
    \begin{minipage}[t]{\linewidth}
    \centering
    \includegraphics[width=8 cm]{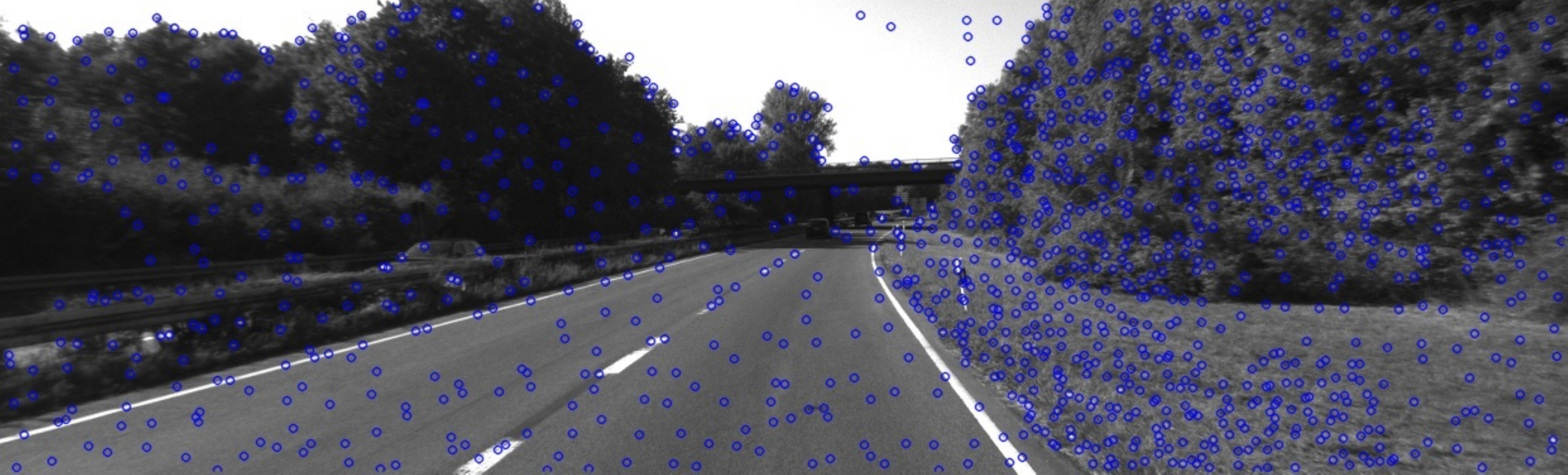}
    \end{minipage}%
    }%
    \\
    \subfigure[Our FAST keypoints of distribution.]{
    \begin{minipage}[t]{\linewidth}
    \centering
    \includegraphics[width=8 cm]{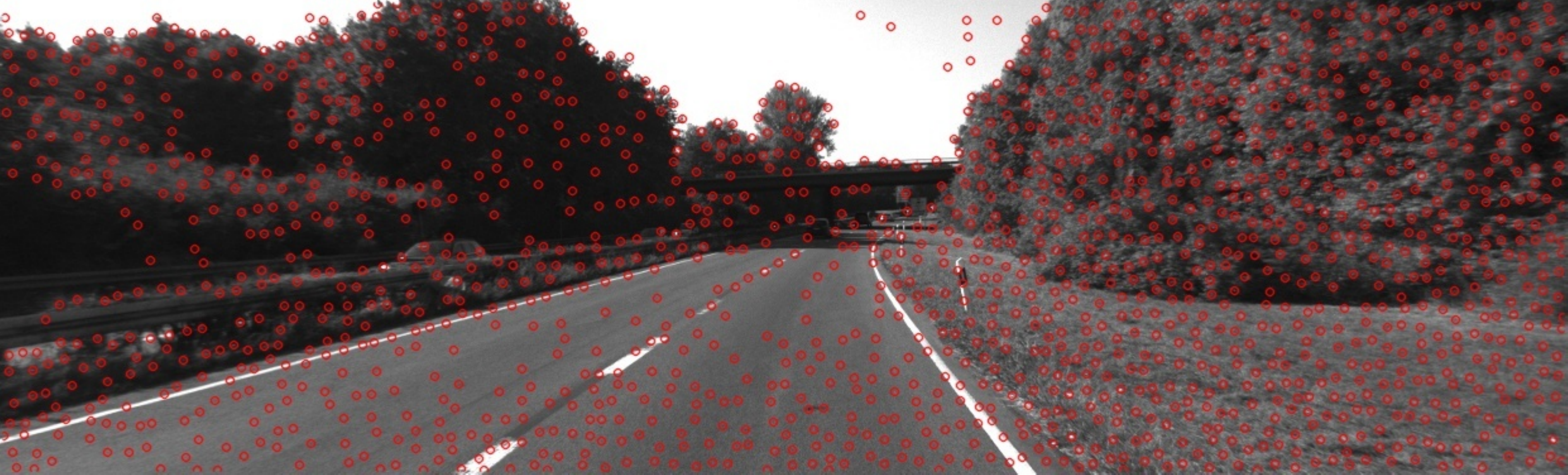}
    \end{minipage}
    }%

\centering
\caption{Comparative results of different distributions for FAST keypoints. Note that in (a), (b), and (c), the green points, the blue points, and the red points indicate the keypoints of position in the frame, respectively.}
\label{img FAST}
\end{figure}

Table \ref{I} shows the time for keyPoints allocation in KITTI and EuRoC datasets. Results show that we obtain the average consumed timeless time to distribute keypoints for each frame than ORB-SLAM2. Fig .\ref{img FAST} show the results for distribution from different methods. Results show that our approach presents a well uniform distribution. 

\begin{table*}[t]
    \centering
    \caption{The time for keyPoints allocation in KITTI and EuRoC datasets. }
    \begin{tabular}{cccccccccccc}
        \hline
        Datasets & MH-01& MH-05 & V1-02 & V1-03 & V2-02 & V2-03& KITTI-00 & KITTI-01 & KITTI-02 & KITTI-03 & KITTI-04\\
        \hline
        ORB-SLAM2(ms) & 23.79 & 22.80 & 25.63 & 24.07 & 23.92 & 66.00 & 25.33 & 25.65 & 16.39 & 25.91 & 25.69\\
        Ours(ms) & 9.20 & 7.01 & 5.92 & 4.82 & 6.76 & 6.27 & 9.29 & 8.87 & 8.68 & 10.08 & 9.00\\
        \hline
    \end{tabular}
    \label{I}
\end{table*}

\subsection{Pose Estimation and Keyframe numbers}
To evaluate our method in different environments, we select the indoor and outdoor datasets, which are EuRoC and KITTI datasets. Note that numbers of lines are $200$ in EuRoC datasets and $500$ in KITTI datasets. Several open-source approaches are compared in this section, including ORB-SLAM2 and Qians' PL-SLAM presented in this paper. We evaluate our method's performance in terms of localization accuracy, computation time, and keyframe numbers. We use $RMSE$(Root Mean Square Error) of $RPE$(Relative Pose Error) and $ATE$(Absolute Trajectory Error) from evo\cite{grupp2017python} tool. "$-$" means that the system failed to run the program in the sequence. \par
\begin{figure}[htbp]

\begin{minipage}{\linewidth}
\centerline{\includegraphics[width=8 cm, trim=2 2 2 2,clip]{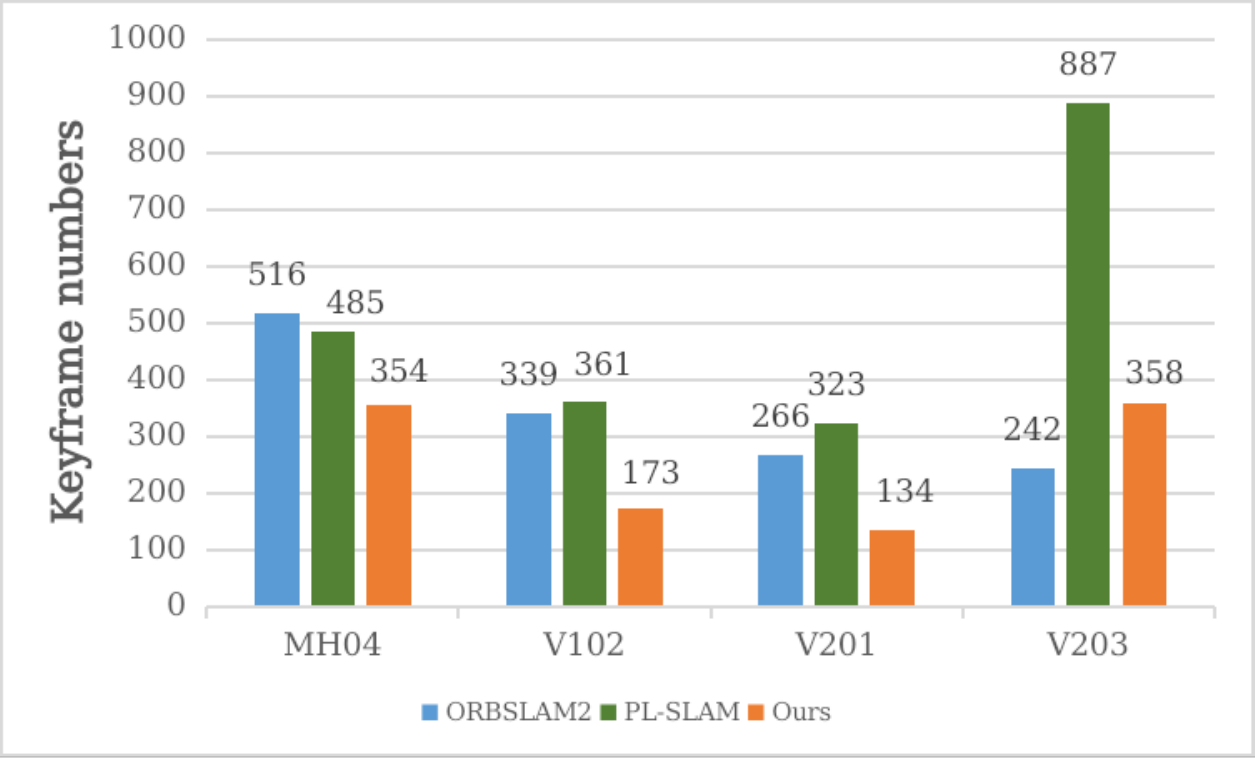}}
\end{minipage}
% \qquad
% \begin{minipage}{0.48\linewidth}
% \centerline{\includegraphics[width=8 cm, trim=2 2 2 2,clip]{times}}

% \end{minipage}

\caption{Keyframe numbers compared with ORB-SLAM2, PL-SLAM, and our method on different sequences on EuRoC.}
\label{1-17-KF}
\end{figure}
\begin{figure}[htbp]
    \subfigure[The processing time of a frame on EuRoC MH-04-difficult.]{
    \begin{minipage}[t]{0.48\linewidth}
    \centering
    \includegraphics[width=4.8 cm]{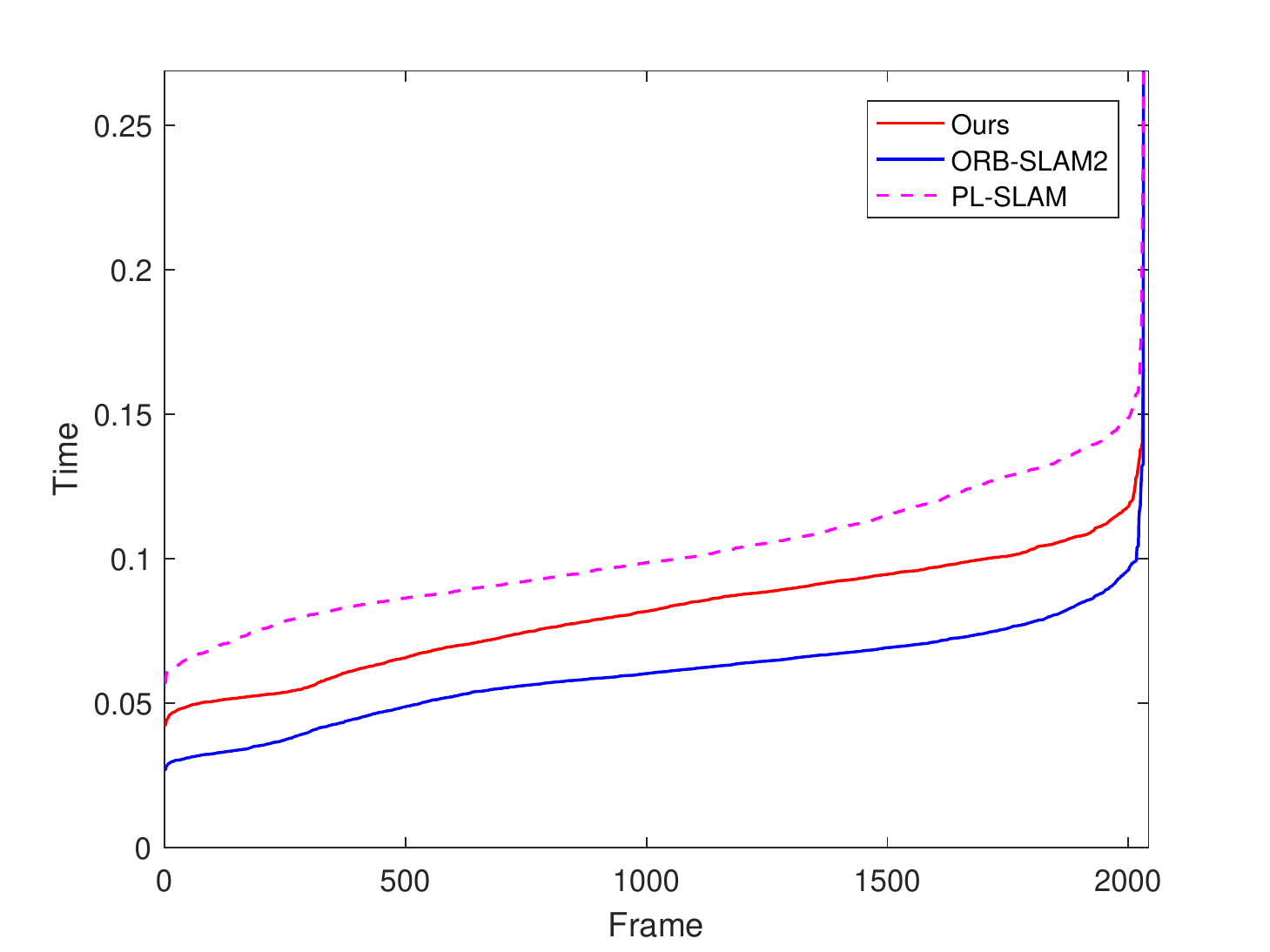}
    \end{minipage}
    }%
    \subfigure[The processing time of a frame on EuRoC V1-02-medium.]{
    \begin{minipage}[t]{0.48\linewidth}
    \centering
    \includegraphics[width=4.8 cm]{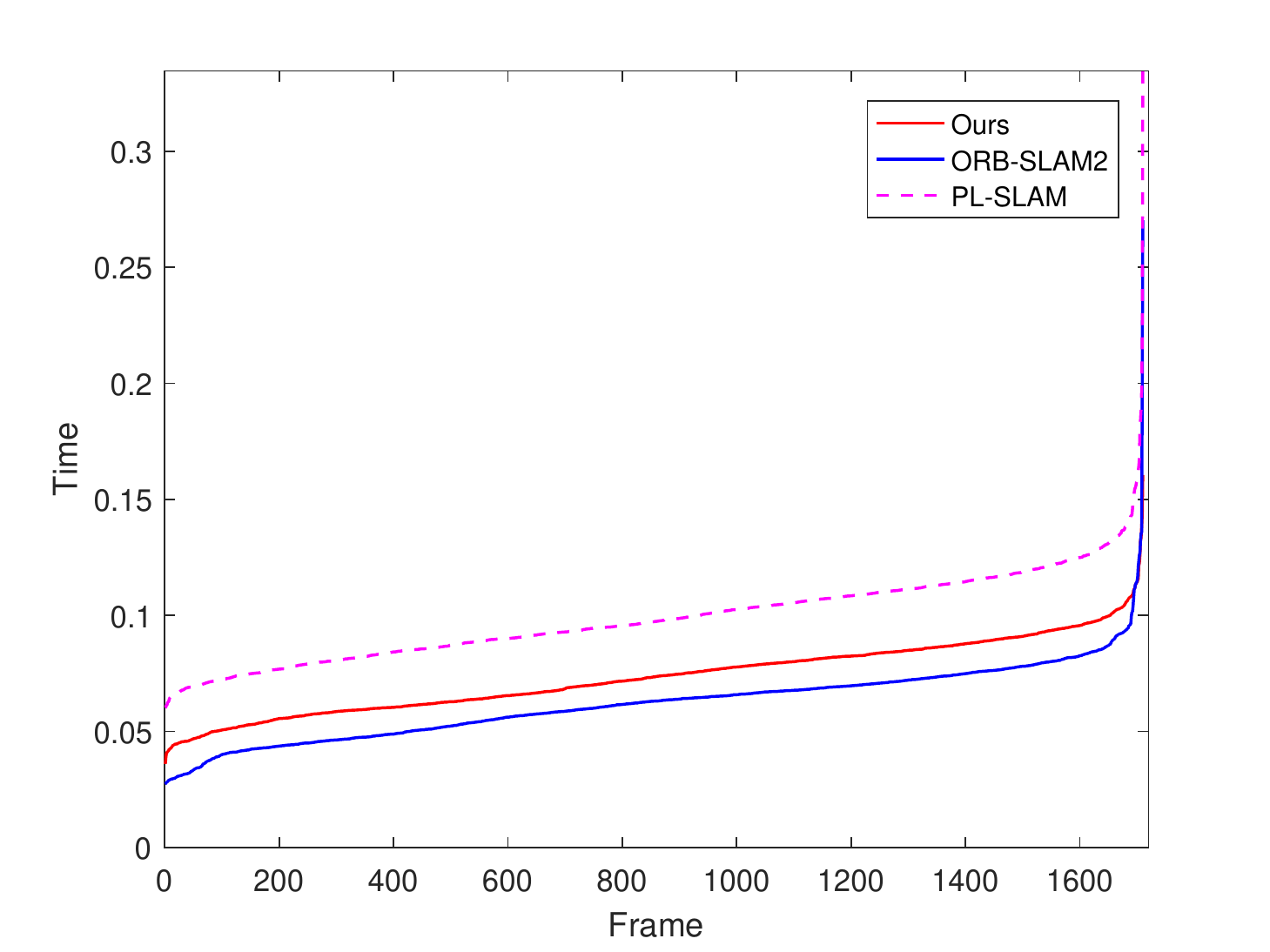}
    \end{minipage}
    }%
    \\
    \subfigure[The processing time of a frame on EuRoC V2-01-easy]{
    \begin{minipage}[t]{0.48\linewidth}
    \centering
    \includegraphics[width=4.8 cm]{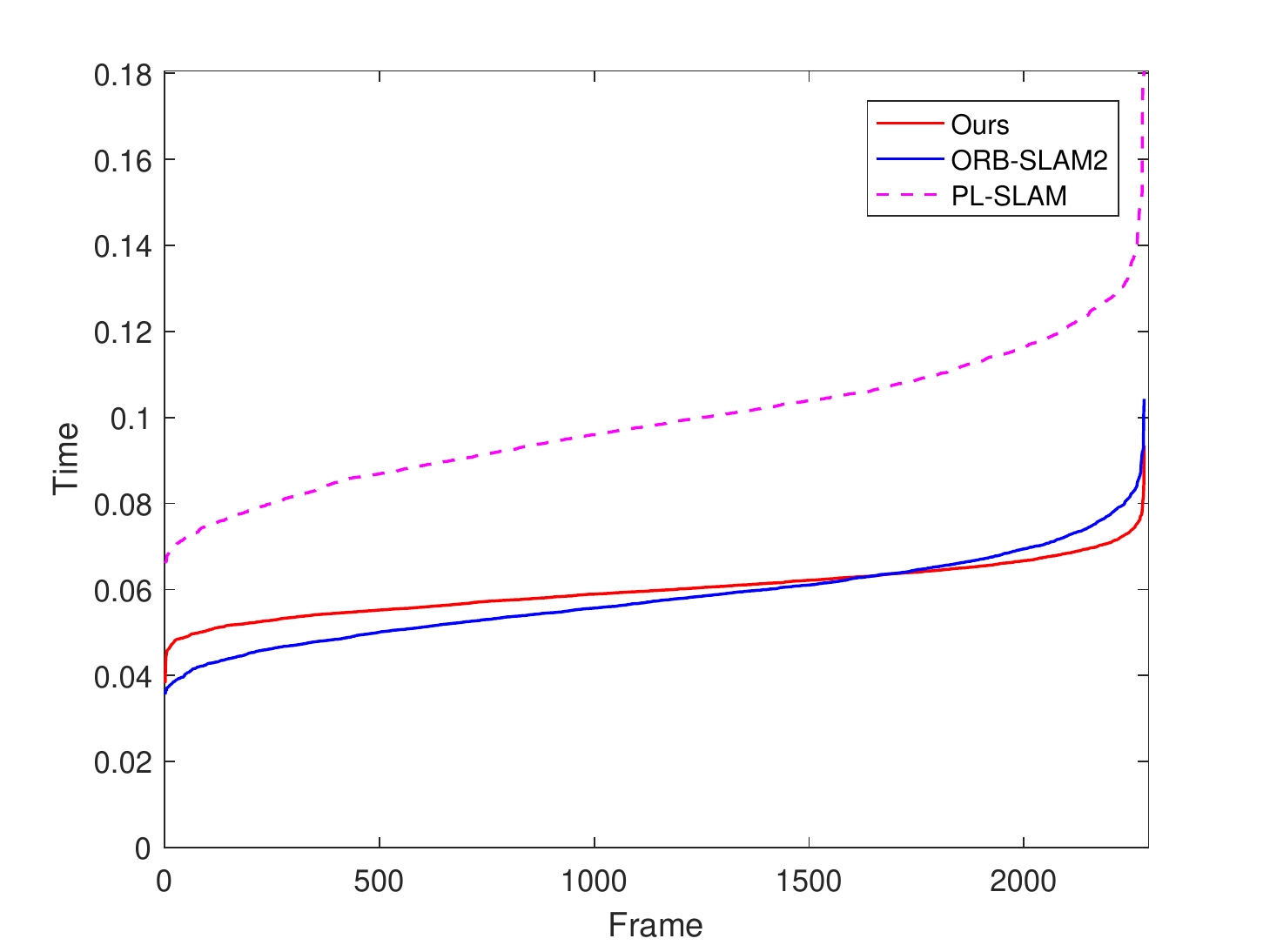}
    \end{minipage}
    }%
    \subfigure[The processing time of a frame on EuRoC V2-03-difficult]{
    \begin{minipage}[t]{0.48\linewidth}
    \centering
    \includegraphics[width=4.8 cm]{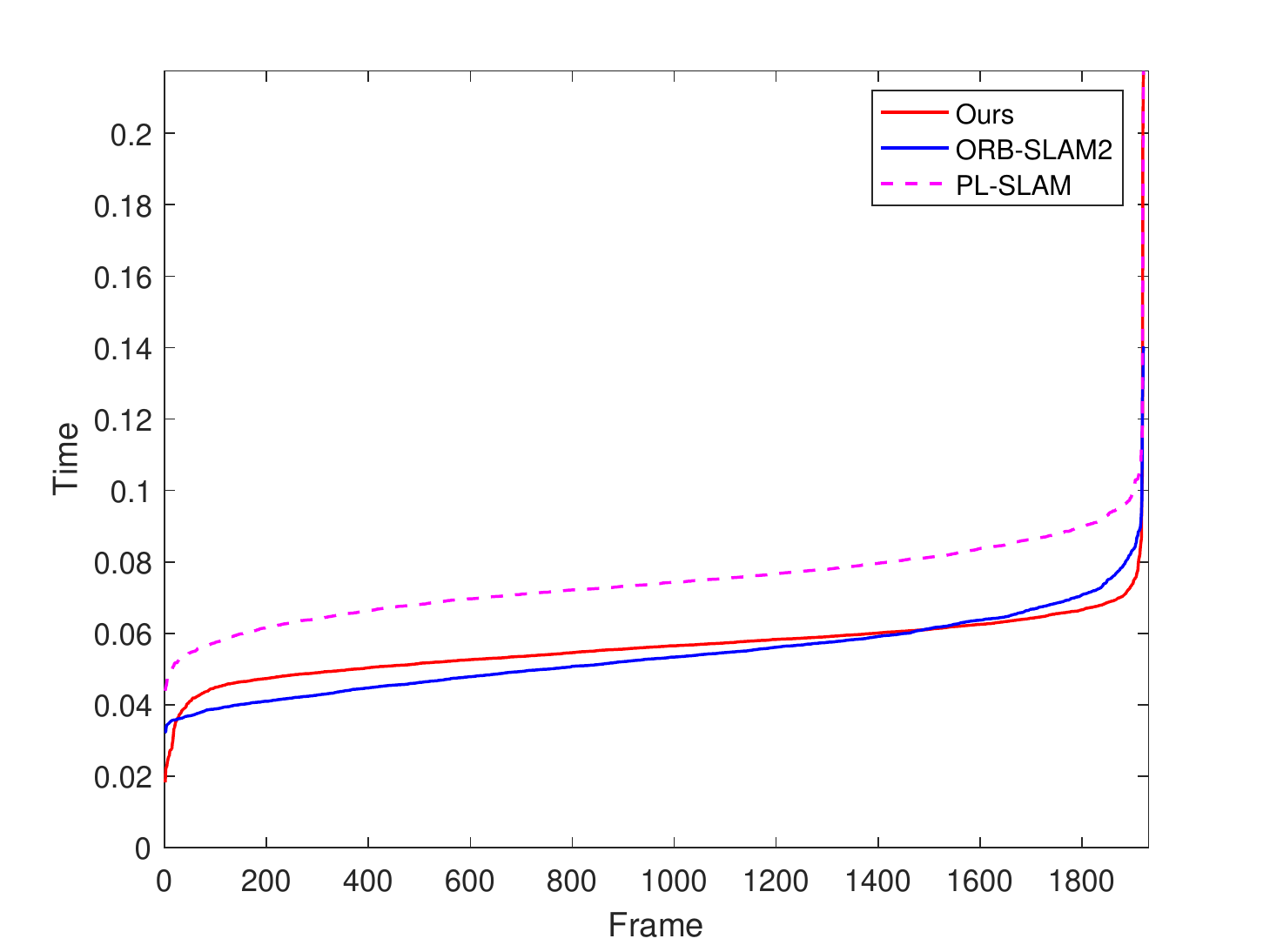}
    \end{minipage}
    }%

\caption{The processing time of a frame on EuRoC. Noted that red is our method, pink is PL-SLAM, and blue is ORB-SLAM for the processing time.}
\label{Time EuRoC}
\end{figure}
\begin{figure*}[htbp]
\begin{minipage}{0.48\linewidth}
\centerline{\includegraphics[width=8 cm, trim=2 2 2 2,clip]{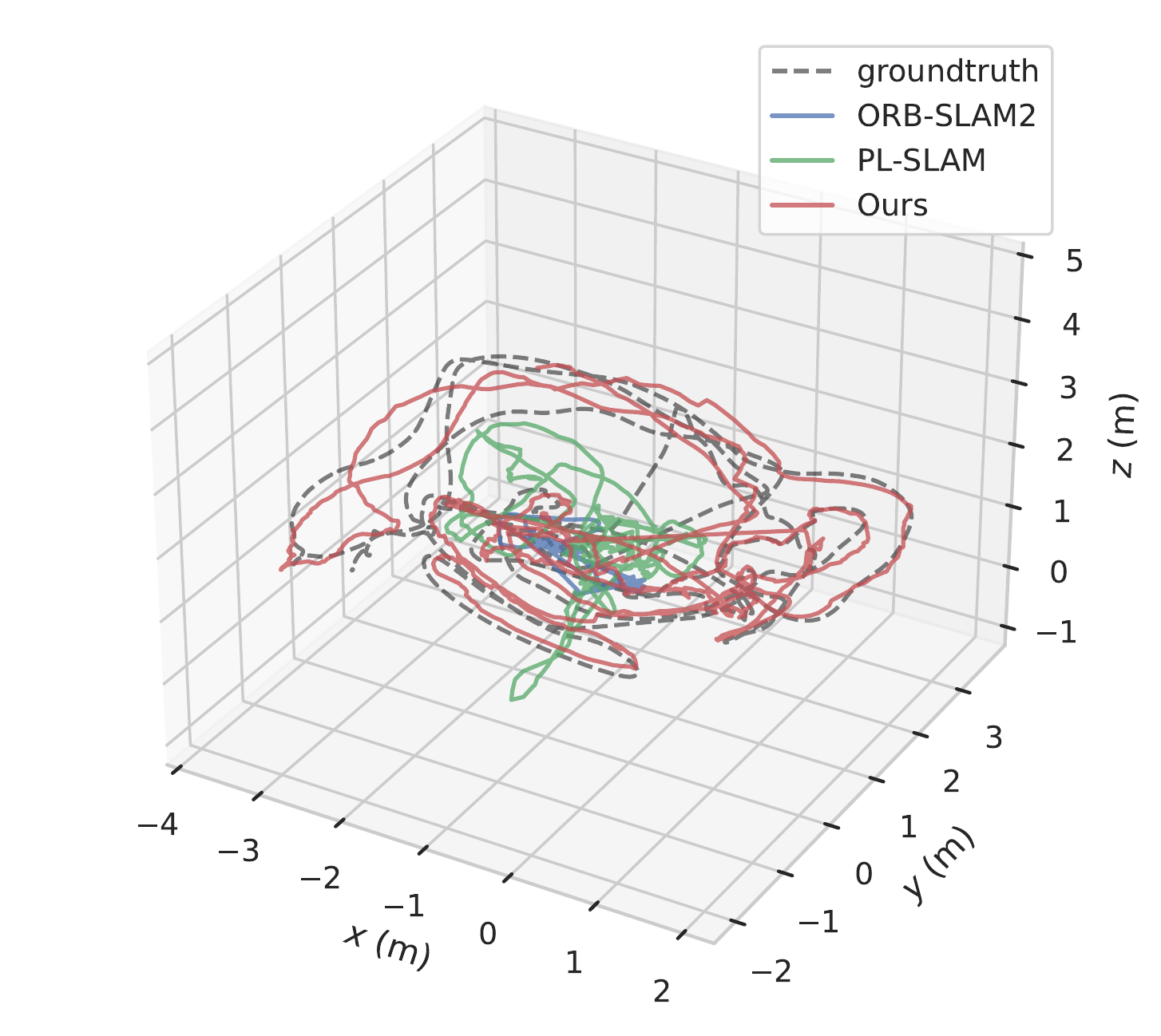}}
\end{minipage}
\qquad
\begin{minipage}{0.48\linewidth}
\centerline{\includegraphics[width=8 cm, trim=2 2 2 2 clip]{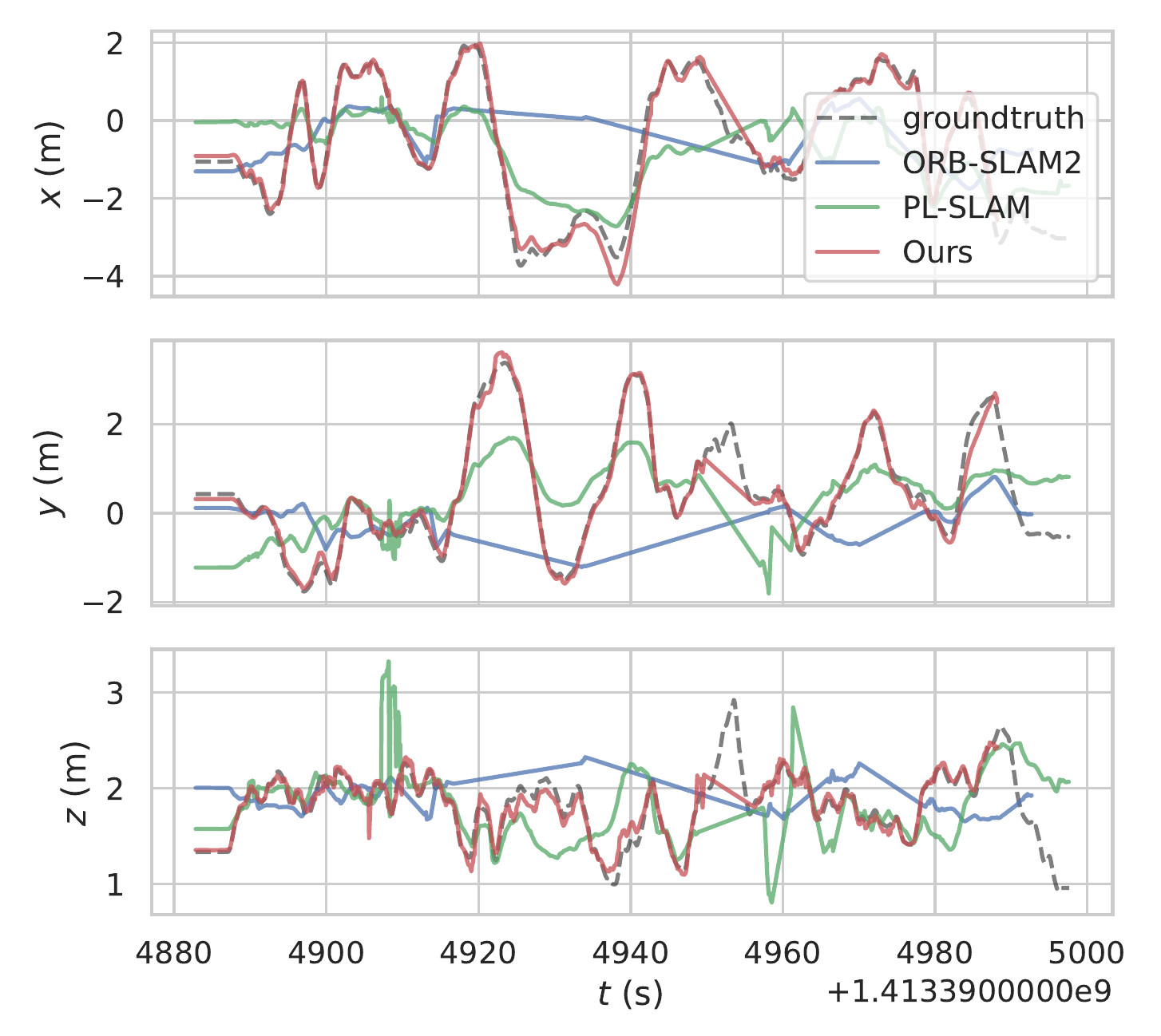}}
\end{minipage}
\caption{Localization accuracy comparison of our method with ORB-SLAM2 and PL-SLAM. The left results show the 3D motion trajectories, and the right results show trajectories with 2D. }
\label{203}
\end{figure*}

\noindent
\textbf{EuRoC datasets:}All $7$ sequences of EuRoC dataset were tested. Results are presented in Table \ref{II} and Fig .\ref{1-17-KF}, respectively. Table \ref{II} shows the experiment results, where $Trans$ and $Rot$ represent $RPE$ of the translations and rotations, respectively. $ATE$ represents $APE$ of the translations. The $RMSE$ (Table \ref{II}) shows that ORB-SLAM2 is better than other methods in low drift error in V102-medium and V201-easy to used FAST of low-threshold. Although PL-SLAM has extract lines in a low texture environment, FAST points of low-threshold lead to large drift. Meanwhile, PL-SLAM has a huge time consumption in Fig .\ref{Time EuRoC}. In addition, PL-SLAM has huge keyframes to finish the program as Fig .\ref{1-17-KF}. Huge keyframes indicate low tracking points and lines. Huge keyframes also indicate huge computing power and resources. Our robust method has low drift in $ATE$. At the same time, our method performs well in $RPE$ of rotation and translation as Table \ref{II}. ORB-SLAM2 can not extract more texture. Compared to our method, it performs poorly in complex datasets. ORB-SLAM2 runs in V203-difficult as Fig .\ref{203} at 14s and 34s. ORB-SLAM2 has a tracking loss. But PL-SLAM and ours can be successfully tracked.\ref{203} and shown ORB-SLAM2 has poor performance. Our method is more precise. We also demonstrated Fig .\ref{Time EuRoC}, comparing with time, our method has less time to finish the program than ORB-SLAM2. This method performs better than ORB-SLAM2 2 and PL-SLAM in low texture indoor environments. We have less time to locate and map. In addition, we have higher localization accuracy and smaller keyframes show that our tracking method is robust.  
\begin{table*}[t]
    \centering
    \caption{Results of ORB-SLAM2, PL-SLAM, and Ours on EuRoC datasets}
    \begin{tabular}{c|ccc|ccc|ccc}
        \hline
         & 
        \multicolumn{3}{c|}{ORB-SLAM2} 
        &\multicolumn{3}{c|}{PL-SLAM}
        &\multicolumn{3}{c}{Ours}\\
        \hline
        datasets& ATE(m) & Rot(rad) & Trans(m) & ATE(m) & Rot(rad) & Trans(m) & ATE(m) & Rot(rad) & Trans(m)\\
        \hline
        MH-04-difficult   & 0.46  & 0.0166   & 0.23           & 0.67      &    0.020 &    0.083       & \textbf{0.21}  & \textbf{0.0112}           &       \textbf{0.069}\\
        MH-05-difficult   & 0.30  & 0.017    & 0.075         & -          &        - &              - & \textbf{0.07}  & \textbf{0.014}  & \textbf{0.061}\\
        V1-02-medium   & \textbf{0.064}  & 0.042     & 0.053          & 0.064        & 0.041 &        0.053    & 0.067          & 0.041     & 0.053\\
        V1-03-difficult   & 0.13     & 0.05  & \textbf{0.048}        & -              &        -        &        -        &\textbf{0.11}  & 0.05     & 0.049\\
        V2-01-easy   & \textbf{0.06}  & 0.02      & 0.019 & 0.082       &        0.02    &        0.02     & 0.076          & \textbf{0.0185}     & 0. 019\\
        V2-02-medium   & 0.30  & 0.046     & 0.048          & -         &        -    &        -     & \textbf{0.11}   & \textbf{0.043} & \textbf{0.042}\\
        V2-03-difficult & 1.25  & 0.18     & 0.35           & 1.66      &        0.11 &        0.17  & \textbf{0.56}   & \textbf{0.079}   & \textbf{0.12}\\
        \hline
    \end{tabular}
    \label{II}
\end{table*}

\noindent
\textbf{KITTI datasets: }Experiments are conducted on all five sequences of KITTI datasets. The KITTI datasets provide large-scale outdoor sequences. We run each sequence three times on our method to have a precision relust. \par
\begin{figure}[htbp]
\begin{minipage}{\linewidth}
\centerline{\includegraphics[width=8 cm, trim=2 2 2 2,clip]{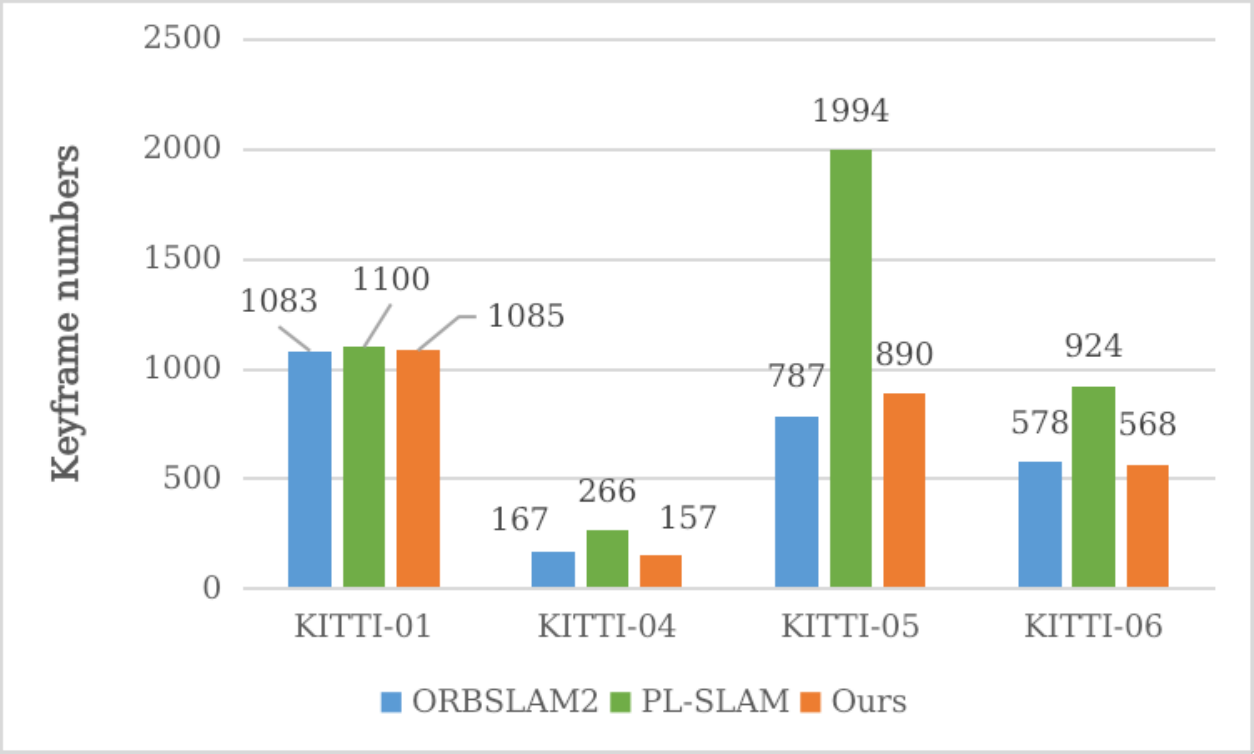}}
\end{minipage}

\caption{Keyframe numbers in comparison with ORB-SLAM2, PL-SLAM, and our method on different sequences(left) on KITTI and a comparison of average runtime for one of the input frames(right) on KITTI. }
\label{Kitti kf time}

\end{figure}
\begin{figure}[htbp]
    \subfigure[The processing time of a frame on KITTI-01.]{
    \begin{minipage}[t]{0.48\linewidth}
    \centering
    \includegraphics[width=4.8 cm]{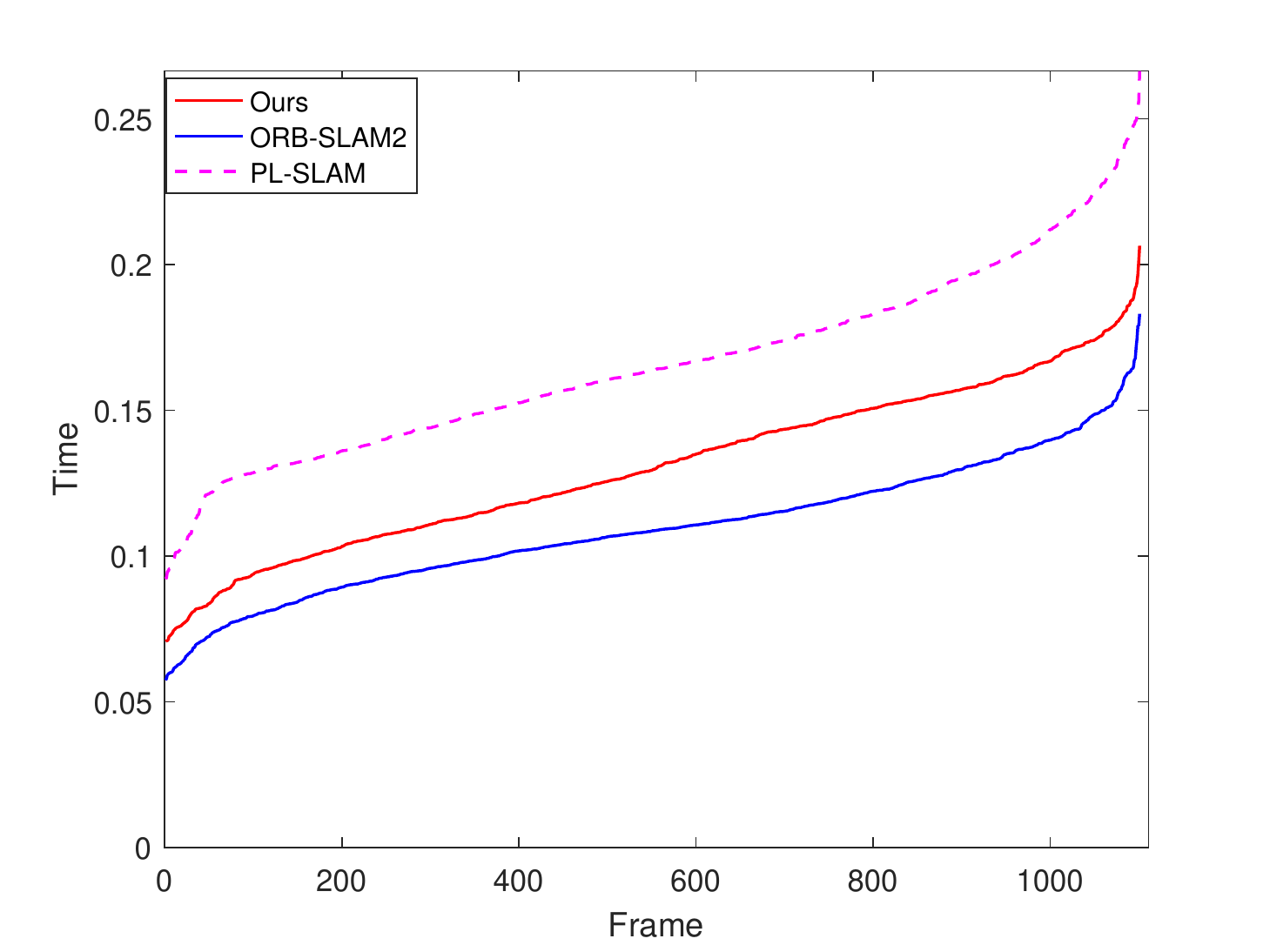}
    \end{minipage}
    }%
    \subfigure[The processing time of a frame on KITTI-04.]{
    \begin{minipage}[t]{0.48\linewidth}
    \centering
    \includegraphics[width=4.8 cm]{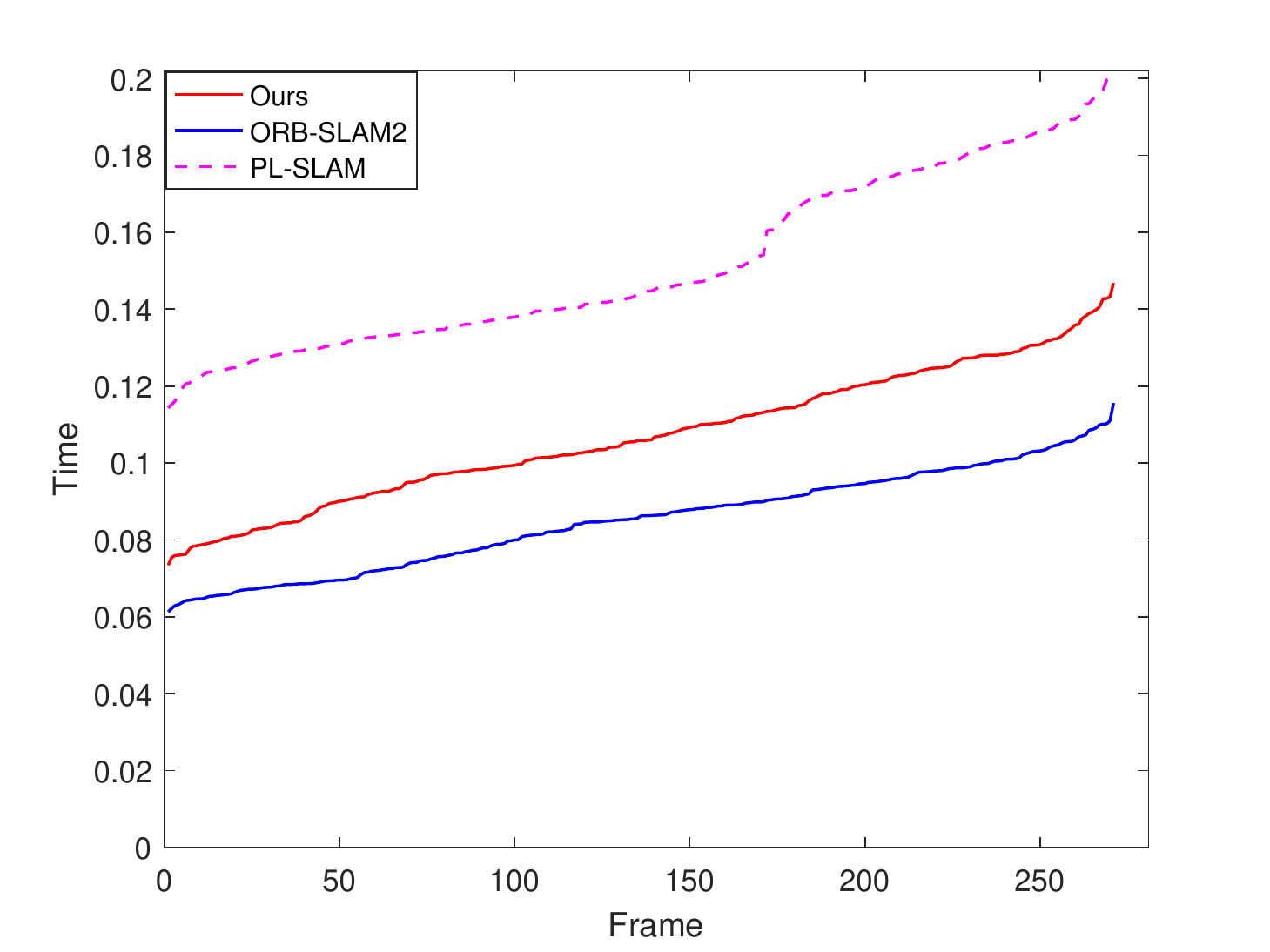}
    \end{minipage}
    }%
    \\
    \subfigure[The processing time of a frame on KITTI-05.]{
    \begin{minipage}[t]{0.48\linewidth}
    \centering
    \includegraphics[width=4.8 cm]{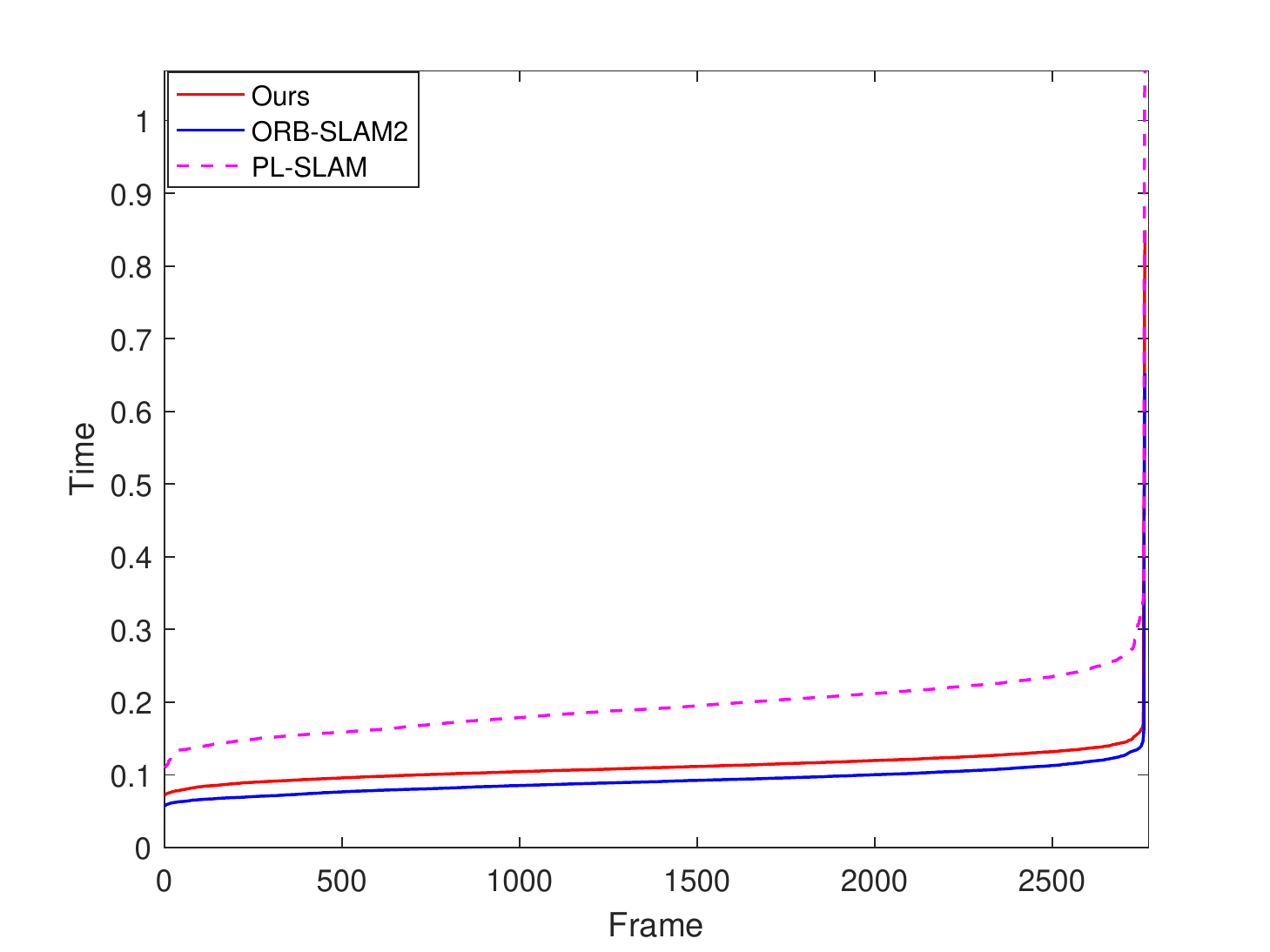}
    \end{minipage}
    }%
    \subfigure[The processing time of a frame on KITTI-06.]{
    \begin{minipage}[t]{0.48\linewidth}
    \centering
    \includegraphics[width=4.8 cm]{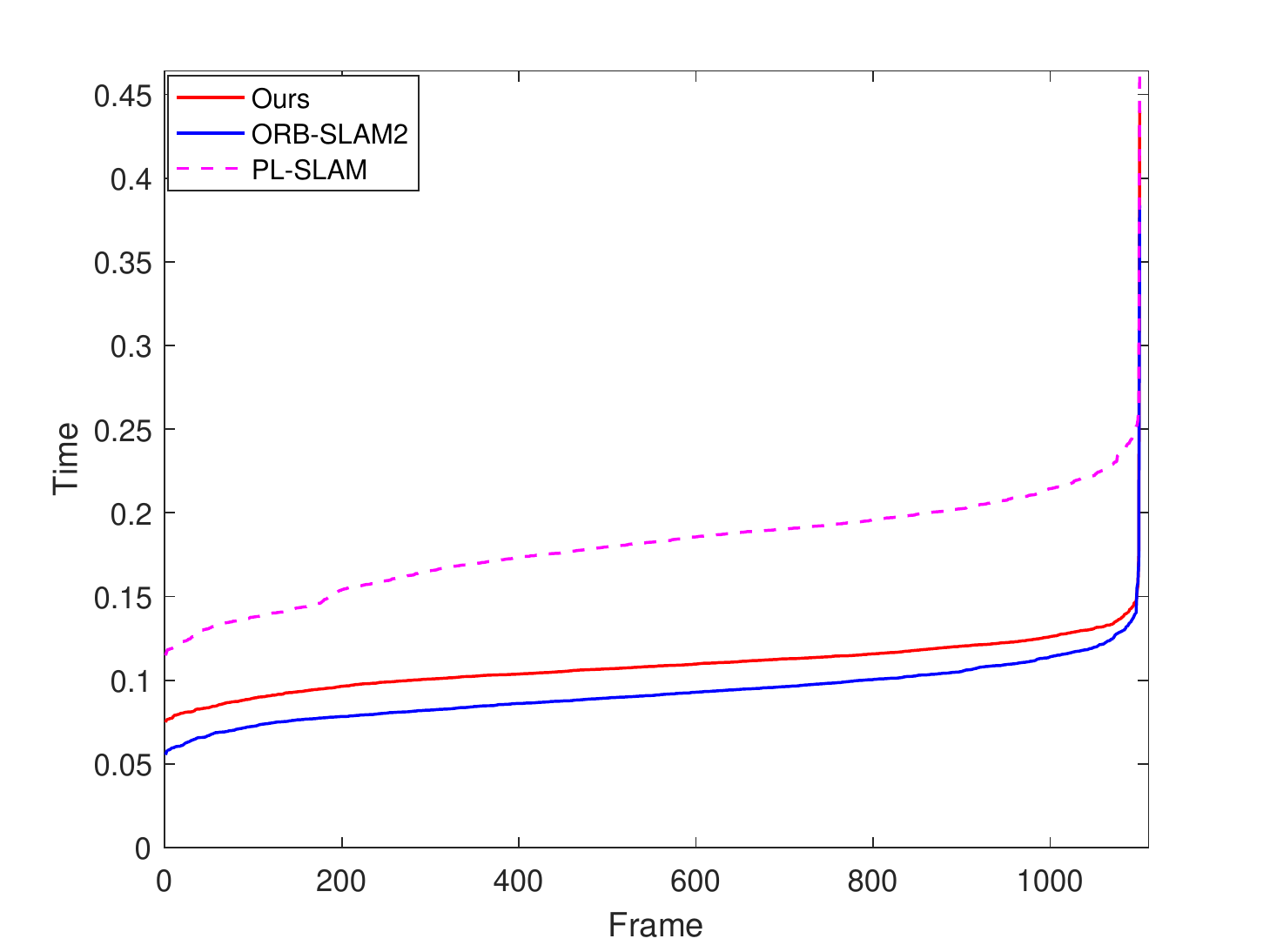}
    \end{minipage}
    }%

\caption{The processing time of a frame on KITTI. Noted that red is our method, pink is PL-SLAM, and blue is ORB-SLAM for the processing time.}
\label{Time KITTI}
\end{figure}

Table \ref{III} shows the results of $RMSE$ in the experiment, where $Trans$ and $Rot$ represent $RPE$ of the translations and rotations, respectively. $ATE$ represents $APE$ of the translations. Table \ref{III} shows that only points SLAM like ORB-SLAM2 will lead to error accumulation. But SLAM has lines that will decrease error. PL-SLAM gets a low drift error as Table \ref{III}. However, PL-SLAM gets the enormous time that it can not run in
real-time, as Fig .\ref{Kitti kf time}. 
% Our method in KITTI-05 has a huge drift due to our approach having Global Bundle Adjustment without the line.
Our method compared with PL-SLAM, our method track landmark more steady as Fig .\ref{Kitti kf time}. In Fig .\ref{Time KITTI}, our approach takes less time to locate. ORB-SLAM2 has poor performance compared with PL-SLAM and Ours \par
In Table \ref{III}, in KITTI-01 datasets, estimation of cameras location has drift due to (\ref{1-17}). One point in the world coordinate system is $P_w$ and two frames observe this point in the camera coordinate system named $P_{C1}(X_{C1}, Y_{C1}, Z_{C1})$ and $P_{C2}(X_{C1}, Y_{C1}, Z_{C1})$, respectively. In the camera coordinate $C1$, the camera coordinate $C2$ and the word coordinate $W$, they have rotation \textbf{$R$} and translation \textbf{$t$} for them. Because of far points, the program ignores \textbf{$t$}. In order to reduce drift, we propose a Kalman algorithm to estimate velocity between referenced Keyframe and the current frame. The camera's velocity can be greater than $7 m/s$ will add keyframes. This way, we can do multiple estimates to reduce ignoring \textbf{$t$}. 
\begin{equation}\label{1-17}
  \begin{bmatrix}
  X_{C1} \\
  Y_{C1}\\
  Z_{C1}
  \end{bmatrix}
  = \textbf{R}\begin{bmatrix}
  X_{C2}\\
  Y_{C2}\\
  Z_{C2}
  \end{bmatrix}
  +\textbf{t}. 
\end{equation}
\par
\begin{figure*}[htbp]
\begin{minipage}{0.3\linewidth}
\centerline{\includegraphics[width=6 cm, trim=2 2 2 2,clip]{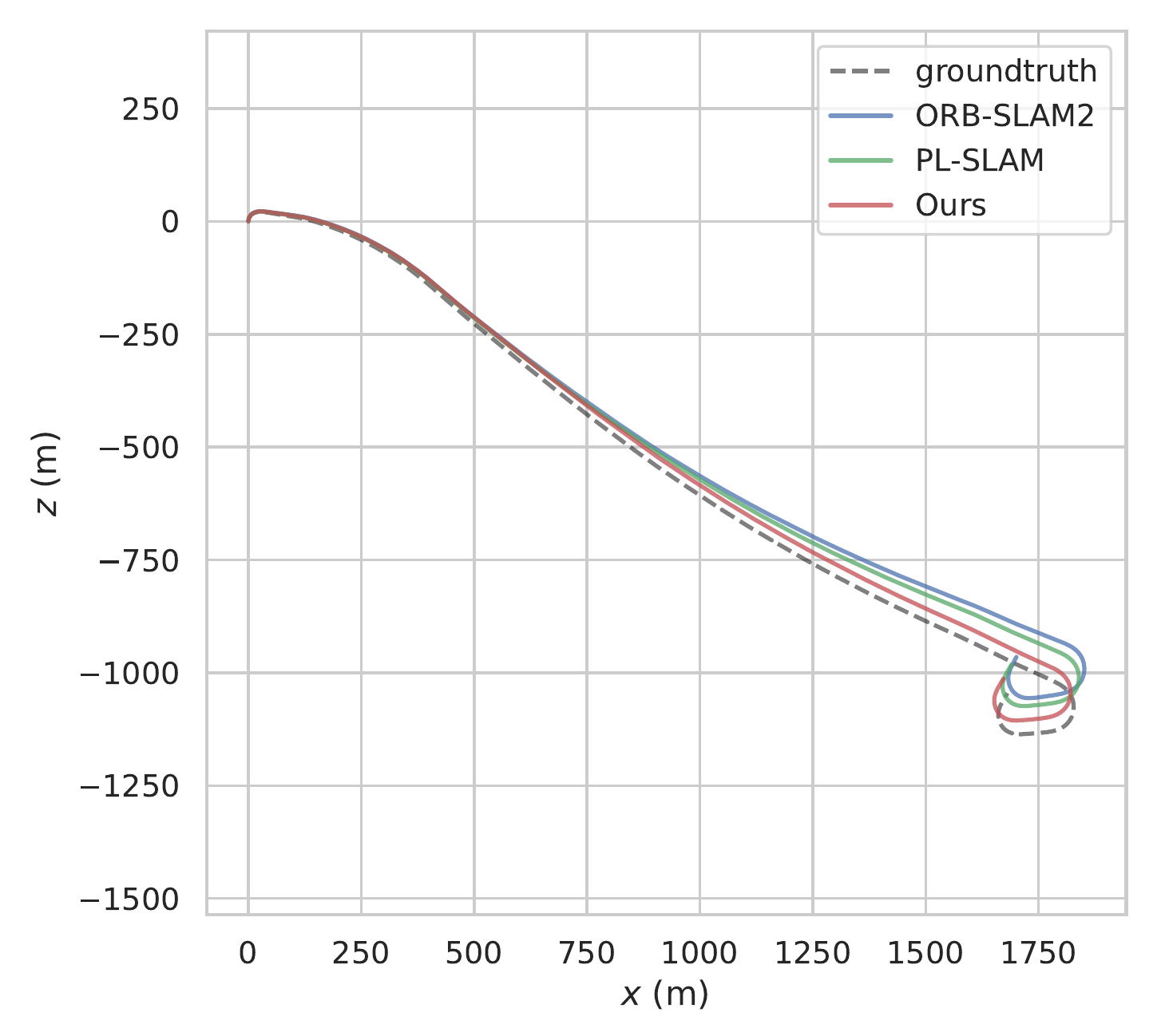}}
\end{minipage}
\qquad
\begin{minipage}{0.3\linewidth}
\centerline{\includegraphics[width=6 cm, trim=2 2 2 2,clip]{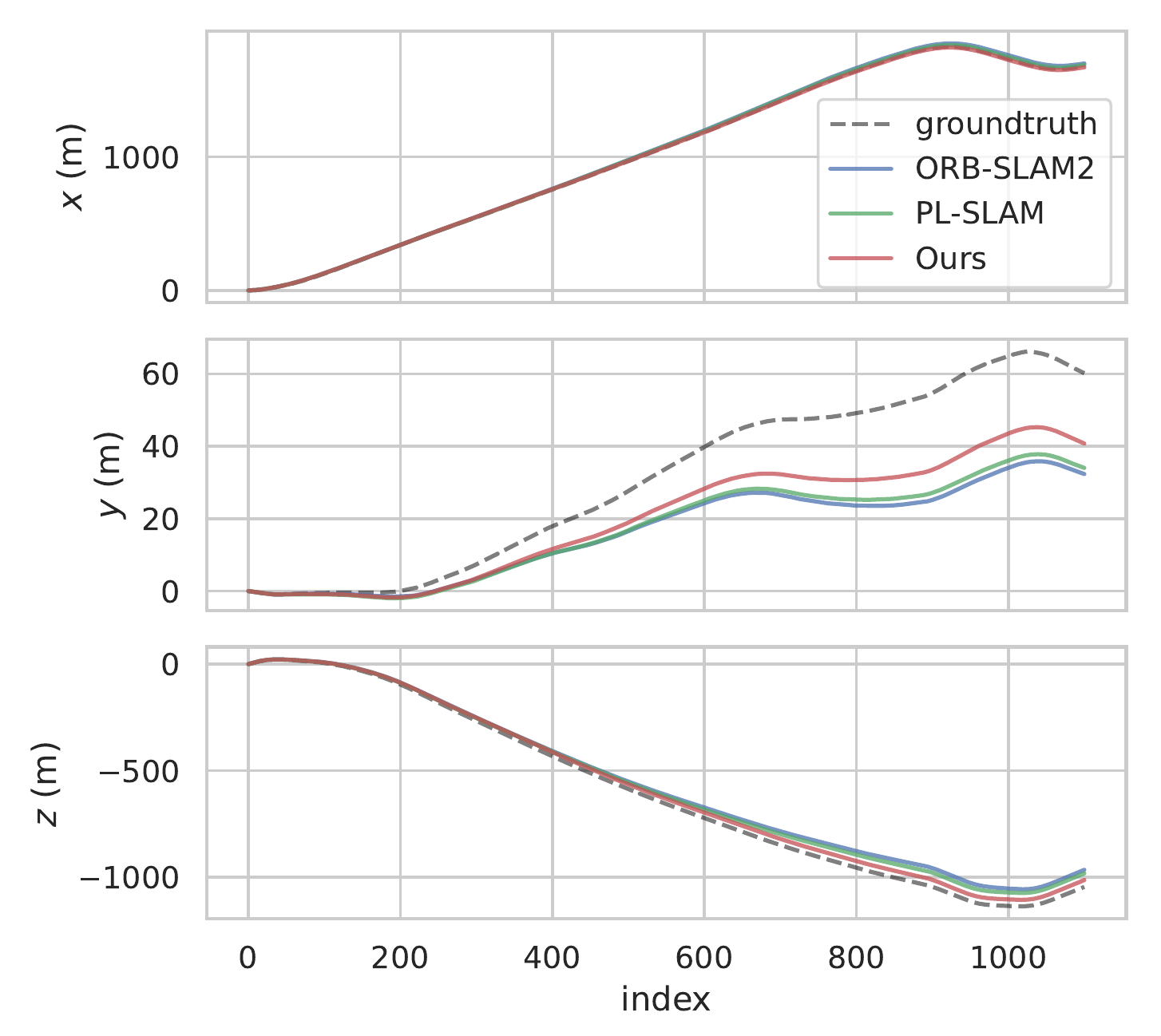}}
\end{minipage}
\qquad
\begin{minipage}{0.3\linewidth}
\centerline{\includegraphics[width=6 cm, trim=2 2 2 2,clip]{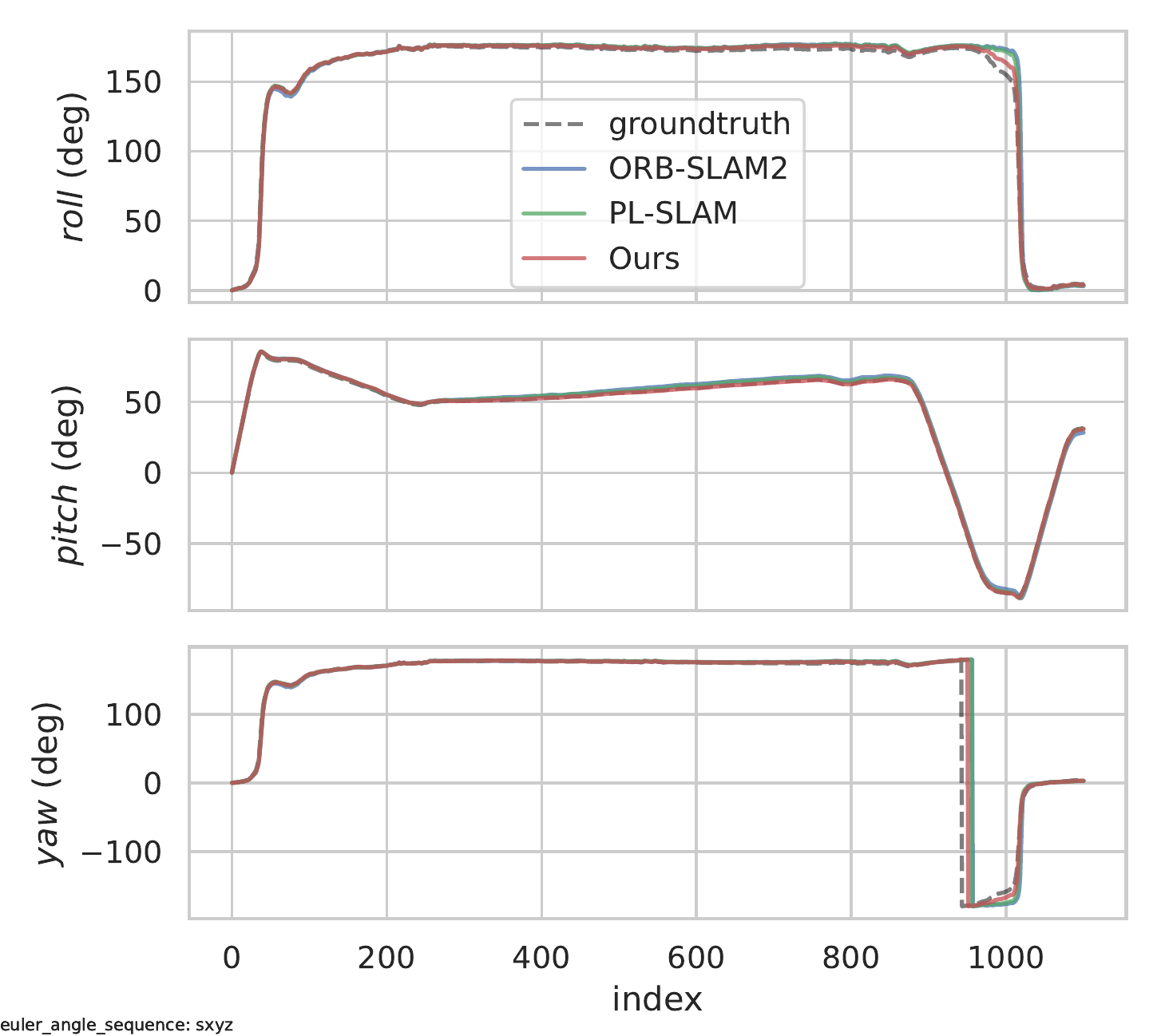}}
\end{minipage}
\caption{Localization accuracy comparison of our method with ORB-SLAM2 and PL-SLAM without $SE(3)$ $Umeyama$. Left cols of results show 3D trajectories on xz coordinate system, the second results show translations of trajectories, and the last results show the rotation of trajectories. }
\label{kitti all}
\end{figure*}
\begin{figure}
  \centering
  \includegraphics[width=8.5cm, trim=2 2 2 2,clip]{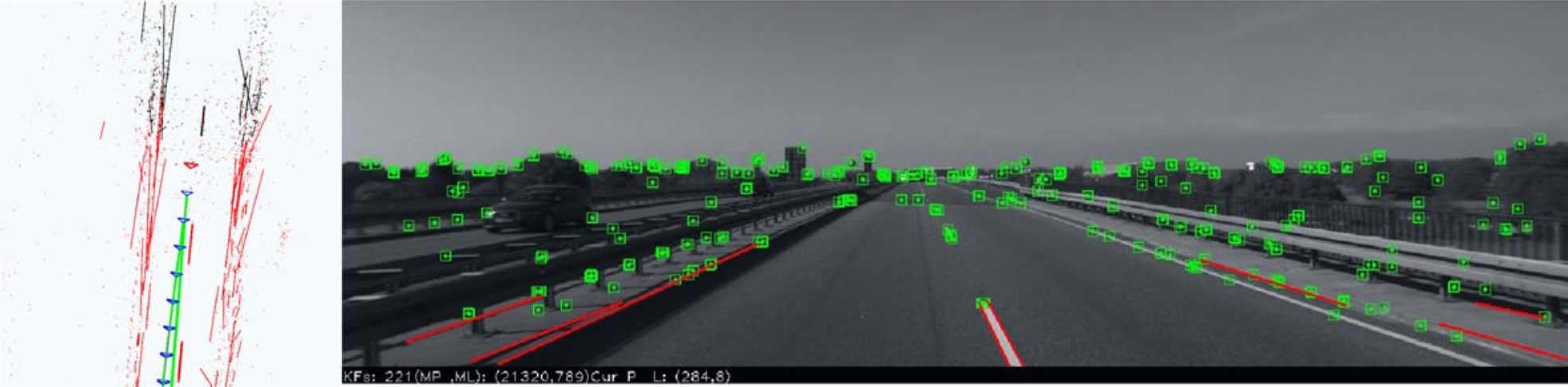}\\
 
  \caption{The proposed visual SLAM with point and line features on our
program on KITTI-01.}
 \label{kitti map img}
\end{figure}

In Fig .\ref{kitti all}, the result shows our method has localization accuracy greater than ORB-SLAM2 and PL-SLAM. ORB-SLAM2 use far points to estimate state, leading to a vast error translation of around $5. 56m$ on $ATE$. PL-SLAM tries to use lines to reduce error translation but shows the extensive time as Fig .\ref{Kitti kf time}. Our method solves the problem with lines and keyframes between the Kalman algorithm and $PID$ leading to less time and fewer keyframes as Fig\ref{Kitti kf time}. Fig\ref{kitti map img} shows our system's effect on the KITTI-01 dataset
. Our system connects disconnected segments to create a better mapping than PL-SLAM. 
\begin{table*}[t]
    \centering
    \caption{Results of ORB-SLAM2, PL-SLAM, and Ours on KITTI datasets.}
    \begin{tabular}{c|ccc|ccc|ccc}
        \hline
         & 
        \multicolumn{3}{c|}{ORB-SLAM2} 
        &\multicolumn{3}{c|}{PL-SLAM}
        &\multicolumn{3}{c}{Ours}\\
        \hline
        datasets& ATE(m) & Rot(rad) & Trans(m) & ATE(m) & Rot(rad) & Trans(m) & ATE(m) & Rot(rad) & Trans(m)\\
        \hline
        KITTI-01   & 5.56  & 0.0009          & 0.049          & 3.73            &0.021             & 0.05          & \textbf{3.21}    & \textbf{0.0007}  & \textbf{0.046}\\
        KITTI-03   & 0.42  & \textbf{0. 0009} & 0.017          & -               & -                & -             & \textbf{0.31}    & 0.001            & 0.017\\
        KITTI-04   & 0.18  & 0.0007          & 0.02           & \textbf{0.16}   &\textbf{0.0006}   &\textbf{0.02}  & 0.21             & 0.0009           & 0.022\\
        KITTI-05   & 0.42  & 0.001           & 0.016          & \textbf{0.38}   &\textbf{0.0009}   &\textbf{0.016} & 0.87             & 0.002            & 0.024\\
        KITTI-06   & 0.90  & 0.0008          & 0.033          & 0.82            &\textbf{0.0007}   &0.018          & \textbf{0.73}    & 0.0009           & \textbf{0.017}\\
         \hline
        
    \end{tabular}
    \label{III}

\end{table*}

\section{Conclusions}
To improve the accuracy and robustness of visual SLAM, we present a graph-based approach using spatial suppression of feature points and line
features with baseline and robust keyframe on $PID$ and Kalman algorithm. Suppression of feature points takes less time to finish the system in a lower threshold value on FAST. The Jacobians of re-projection error concerning the line parameters consider baseline making a good performance. And $PID$ keyframe and Kalman keyframe wise decisions to insert keyframes. They demonstrated that the fusion baseline lines produces more robust estimations in real-world scenarios. In the future, 
we will study how to introduce inertial sensors into our system with point and line features. 

In view of (\ref{1-14}), the Jacobian of the re-projection error with line parameters in detail show at (\ref{1-24}). 
\begin{table*}[t]
\begin{equation}\label{1-24}
  j_{\zeta}=
\begin{bmatrix}
  \frac{f_xl_x}{z\sqrt{l_x^2 + l_y^2}} & \frac{f_yl_y}{z\sqrt{l_x^2 + l_y^2}} &- \frac{f_xl_x+f_yl_yy}{z^2\sqrt{l_x^2 + l_y^2}} & - \frac{f_xl_xy-f_yl_yy^2}{z^2\sqrt{l_x^2 + l_y^2}}-\frac{f_yl_y}{\sqrt{l_x^2 + l_y^2}}&\frac{xf_xl_x+f_yl_yxy}{z^2\sqrt{l_x^2 + l_y^2}}
  + \frac{f_xl_x}{\sqrt{l_x^2 + l_y^2}}& \frac{f_yl_yx-f_xl_xy}{z\sqrt{l_x^2 + l_y^2}}\\
  0 & \frac{f_yl_y}{z\sqrt{lx^2 + ly^2}} & - \frac{bfl_x-f_yl_yy}{z^2\sqrt{l_x^2 + l_y^2}} &- \frac{ybfl_x+f_yl_yy^2}{z^2\sqrt{lx^2 + ly^2)}} - \frac{f_yl_y}{\sqrt{lx^2 + ly^2}} &\frac{xbfl_x+f_yl_yxy}{z^2\sqrt{l_x^2 + l_y^2}} &\frac{f_yl_yx}{z\sqrt{l_x^2 + l_y^2}}
  \\
  0 & 0 & 0 & 0 & 0 & 0\\
  \end{bmatrix} 
\end{equation}
\end{table*}

\bibliographystyle{ieeetr}
\bibliography{main.bib}
\end{document}